%% file: arxiv.tex
\documentclass[11pt]{article}

\date{}
\usepackage{times}

\usepackage[utf8]{inputenc} 
\usepackage[T1]{fontenc}    
\usepackage{hyperref}       
\usepackage{url}            
\usepackage{booktabs}       

\usepackage{amsfonts}       
\usepackage{amsmath}
\usepackage{amsthm}
\usepackage{amssymb}
\usepackage{mathtools}

\usepackage{nicefrac}       
\usepackage{microtype}      
\usepackage{xspace}
\usepackage{xcolor}
\usepackage{xfrac}

\usepackage{algorithm}
\usepackage[noend]{algpseudocode}

\usepackage[numbers, compress]{natbib}
\usepackage{caption}
\usepackage{enumitem}

\usepackage{graphicx}
\usepackage{float}

\input{define.tex}

\title{Vertical Symbolic Regression via Deep Policy Gradient}
\author{Nan Jiang, Md Nasim, Yexiang Xue\\
Department of Computer Science, Purdue University \\
\texttt{\{jiang631,mnasim, yexiang\}@purdue.edu}}

\begin{document}

\maketitle

\begin{abstract}
\input{tex/0.abstract}
\end{abstract}

\input{tex/1.intro}

\input{tex/2.prelim}
\input{tex/3.pipeline}

\input{tex/4.components}

\input{tex/5.related}

\input{tex/6.exp}

\input{tex/7.conclude}

\section*{Acknowledgments}
This research was supported by NSF grant CCF-1918327 and DOE – Fusion Energy Science grant: DE-SC0024583.

\clearpage
\bibliography{reference}
\bibliographystyle{unsrtnat} 
\newpage

\appendix

\input{tex/10.impelment}

\input{tex/10.expset}
\input{tex/10.expextra}

\end{document}

%% file: define.tex
\usepackage{enumitem}
\usepackage{bm}
\usepackage{amsmath,amssymb,amsfonts}
\usepackage{graphicx}

\usepackage{multicol,multirow}
\usepackage{url}
\usepackage{xcolor}
\usepackage{xfrac}
\usepackage{xspace}
\usepackage{algorithm}
\usepackage[noend]{algpseudocode}
\usepackage{mdframed}
\algrenewcommand\algorithmicrequire{\textbf{Input:}}
\algrenewcommand\algorithmicensure{\textbf{Output:}}

\newcommand{\method}{\textsc{Vsr-Dpg}\xspace}

\newcommand{\diff}{\mathrm{d}}

\newcommand{\cvgp}{\text{VSR-GP}\xspace}

%% file: tex/0.abstract.tex
Vertical Symbolic Regression (VSR) recently has been proposed to expedite the discovery of symbolic equations with many independent variables from experimental data.
VSR reduces the search spaces following the vertical 
discovery path by building from reduced-form equations involving a subset of independent variables to full-fledged ones. 
Proved successful by many symbolic regressors, deep neural networks are expected to further scale up VSR. 
Nevertheless, directly combining VSR with deep neural networks will result in difficulty in passing gradients and other engineering issues.
We propose \underline{\textbf{V}}ertical \underline{\textbf{S}}ymbolic \underline{\textbf{R}}egression using \underline{\textbf{D}}eep \underline{\textbf{P}}olicy \underline{\textbf{G}}radient (\method)  and demonstrate that \method can 
recover ground-truth equations involving multiple input variables, significantly beyond both deep reinforcement learning-based approaches and previous VSR variants. 
Our \method models symbolic regression as a sequential decision-making process, in which equations are built from repeated applications of grammar rules.
The integrated deep model is trained to maximize a policy gradient objective. 
Experimental results demonstrate that our \method significantly outperforms popular baselines in identifying both algebraic equations and ordinary differential equations on a series of benchmarks.

%% file: tex/1.intro.tex
\section{Introduction}
\begin{figure}[!t]
    \centering
    \includegraphics[width=0.59\linewidth]{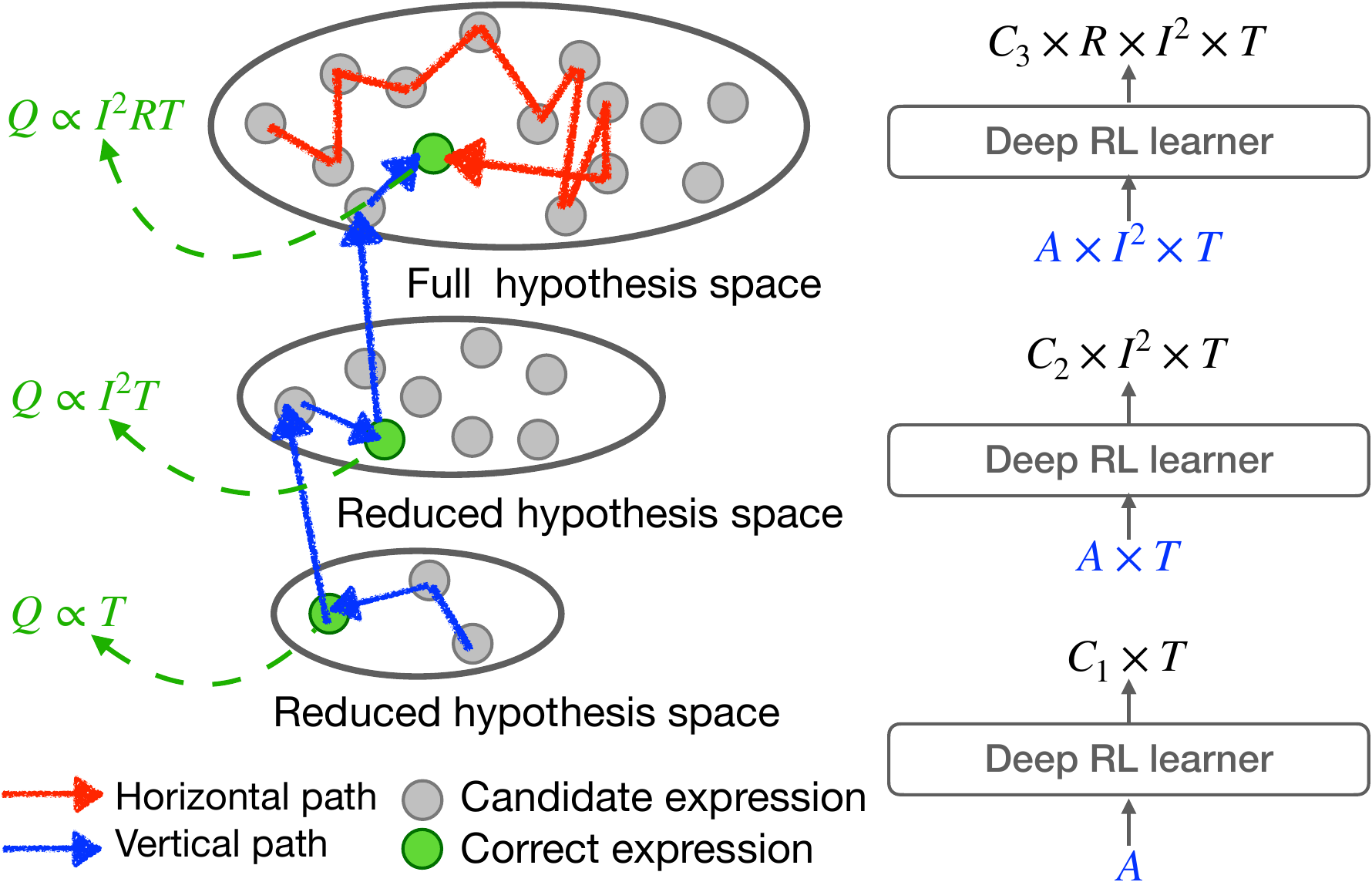}
    \caption{Our \method follows a vertical path (colored blue) better than the horizontal path (colored red), in the scientific discovery of Joule's first law. \textbf{(Left)} The vertical discovery starts by finding the relationship between two factors ($Q, T$) in a reduced hypothesis space with other factors held constant.
    It then finds models in extended hypothesis space with three factors ($Q, I, T$), and finally in the full hypothesis space. 
    Searching following the vertical paths is way cheaper since the sizes of the reduced hypothesis spaces in the first few steps are exponentially smaller than the full hypothesis space. 
    \textbf{(Right)} Our \method extends the equation in each step. The placeholder symbol $A$ indicates a sub-expression.
    } \label{fig:hv}
    \vspace{-.6em}
\end{figure}

Exciting progress has been made to accelerate scientific discovery using Artificial Intelligence (AI)~\cite{langey1988scientificdiscovery,kulkarni1988processes,wang2023scientific}.
Symbolic regression, as an important benchmark task
in AI-driven scientific discovery, 
distills physics models in the form of symbolic equations from experiment data~\cite{doi:10.1126/science.1165893}. 
Notable progress in symbolic regression includes search-based methods~\cite{LENAT1977ubiquity}, genetic programming~\cite{doi:10.1126/science.1165893,DBLP:conf/gecco/VirgolinAB19}, Monte Carlo tree search~\cite{DBLP:conf/icml/TodorovskiD97,DBLP:conf/iclr/Sun0W023,DBLP:conf/icml/KamiennyLLV23}, and deep reinforcement learning~\cite{DBLP:conf/iclr/PetersenLMSKK21,DBLP:conf/nips/MundhenkLGSFP21}.

Vertical Symbolic Regression (VSR) \cite{DBLP:conf/pkdd/JiangX23,jiang2023vertical} recently has been proposed to expedite the discovery 
of symbolic equations with many independent variables. 
Unlike previous approaches, VSR
reduces the search spaces following a vertical 
discovery route -- it extends from reduced-form 
equations involving a subset of independent 
variables to full-fledged ones, adding one 
independent variable into the equation at a time.
Figure~\ref{fig:hv} provides an example. To discover Joule's law $Q\propto I^2RT$, where $Q$ is heat, $I$ is current, $R$ is resistance, and $T$ is time~\cite{joule1843production}, one first holds $I$ and $R$ as constants and finds  $Q \propto T$. 
In the second round, $I$ is introduced into the equation with targeted experiments that study the effect of $I$ on $Q$. Such rounds repeat until all factors are considered.
Compared with the {horizontal routes}, which directly model all the independent variables simultaneously, vertical discovery can be significantly cheaper
because the search spaces of the first few steps 
are exponentially smaller than the full hypothesis space.

Meanwhile, deep learning, especially deep policy gradient~\cite{DBLP:conf/iclr/PetersenLMSKK21,DBLP:conf/nips/MundhenkLGSFP21}, has boosted the performance of symbolic regression approaches to a new level. 
Nevertheless, VSR was implemented using genetic programming. 
Although we hypothesize deep neural nets should boost VSR, it is not straightforward to integrate VSR with deep policy gradient-based approaches. 
The first idea is to employ deep neural nets to predict the symbolic equation tree in each vertical expansion step. 
However, this will result in (1) difficulty passing gradients from trees to deep neural nets and (2) complications concatenating deep networks for predictions in each vertical expansion step. We provide a detailed analysis of the difficulty in Appendix~\ref{apx:direct-integ}.

In this work, we propose \underline{\textbf{V}}ertical \underline{\textbf{S}}ymbolic \underline{\textbf{R}}egression using \underline{\textbf{D}}eep \underline{\textbf{P}}olicy \underline{\textbf{G}}radient (\method). 
We demonstrate that \method can 
recover ground-truth equations involving $50$ variables, which is beyond both deep reinforcement learning-based approaches and previous VSR variants (best up to $6$ variables). 
Our \method solves the above difficulty based on the following key idea: each symbolic expression can be treated as 
repeated applications of grammar rules. 
Hence, discovering the best symbolic equations in the space of all candidate expressions is viewed as the sequential decision-making of predicting the optimal sequence of grammar rules.

In Figure \ref{fig:hv}(right), the expansion from $Q=C_1T$ to $Q=C_2I^2 T$ is viewed as replacing constant $C_1$ with a sub-expression $C_2I^2$. We define a context-free grammar on symbolic expression expansion, denoting the rules that certain constants can be replaced with other variables, constants, or sub-expressions. All candidate expressions that are compatible with $C_1T$ can be generated by repetitively applying the defined grammar rules in different order. \method employs a Recurrent Neural Network (RNN) to sample many sub-expressions (including $C_2I^2$) by sequentially sampling rules from the RNN.  A vertical discovery path is built on top of this sequential decision-making process of reduced-form symbolic expressions.
The RNN needs to be trained to produce expansions that lead to high fitness scores on the dataset. In this regard, we train the RNN to maximize a policy gradient objective, similar to that proposed in~\citeauthor{DBLP:conf/iclr/PetersenLMSKK21}. 
Our biggest difference compared 
with ~\citeauthor{DBLP:conf/iclr/PetersenLMSKK21} is that the RNN in our \method predicts the next rules in the vertical discovery space, while their model predicts the pre-order traversal of the expression tree in the full hypothesis space.

In experiments, we consider several challenging datasets of algebraic equations with multiple input variables and also of real-world differential equations in material science and biology. (1) In Table~\ref{tab:trig}, our \method attains the smallest median NMSE values in 7 out of 8 datasets, against a line of current popular baselines including the original VSR-GP. The main reason is deep networks offer many more parameters than the GP algorithm, which can better adapt to different datasets and sample higher-quality expressions from the deep networks. (2) Further analysis on the best-discovered equation (in Table~\ref{tab:recovery-rate}) shows that \method uncovers up to 50\% of the exact governing equations with $5$ input variables, where the baselines only attain 0\%.
(3)  
In table~\ref{tab:extend-trig}, our \method is able to find high-quality expressions on expressions with more than 50 variables, which is never the case in all the baselines. The major reason is the use of the control variable experiment. (4) On discovery of ordinary differential equations in Table~\ref{tab:diff-equation}, our \method also improves over current baselines.

\begin{figure*}[!t]
    \centering
    \includegraphics[width=0.99\linewidth]{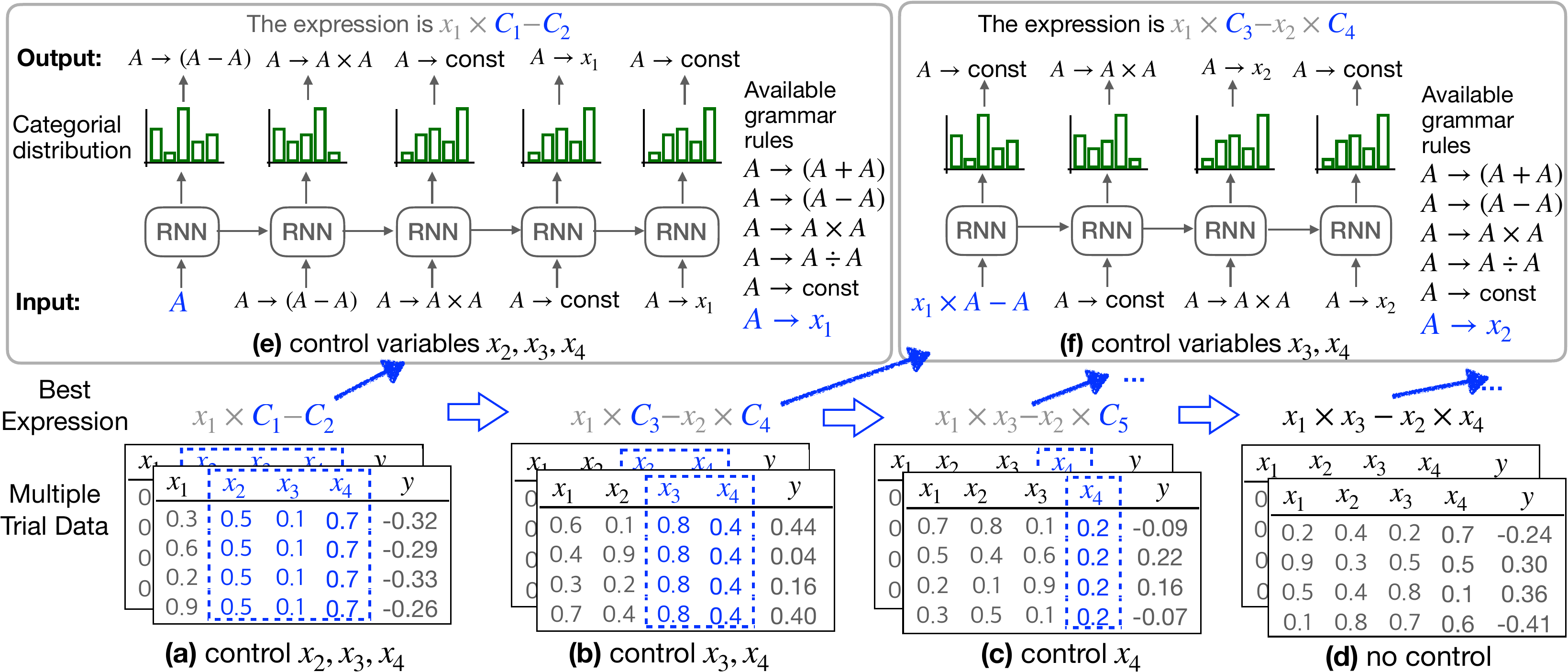}
    \caption{The proposed \method for the discovery of expression $\phi=x_1\times x_3-x_2\times x_4$.  \textbf{(a)} Initially, a reduced-form equation $\phi= x_1\times C_1 - C_2$ is found, in which  $x_2, x_3, x_4$ are held constant and only  $x_1$ is allowed to vary.   $C_1$ and $C_2$ (colored blue) are summary constants, which are sub-expressions containing the controlled variables. 
    The open constants in the expression are fitted by the corresponding controlled variable data.
    \textbf{(b)} In the second stage, this equation is expanded to $x_1\times C_3 - x_2\times C_4$.  \textbf{(c, d)} This process continues until the ground-truth equation $\phi=x_1 x_3 - x_2 x_4$ is found. \textbf{(e, f)} Under those controlled variables, the deep recurrent neural network (RNN) predicts a categorical distribution over the available grammar rules, which only have the free variables (colored blue). The best-predicted expression in (e) is reformulated as the start symbol for in (f) $x_1\times A - A$.}
    \label{fig:motivation}
\end{figure*}

%% file: tex/2.prelim.tex
\section{Preliminaries}

\noindent\textbf{Symbolic Regression}  aims to discover governing equations from the experimental data. An example of such mathematical expression is Joule's first law: $Q=I^2 RT$, which quantifies the amount of heat $Q$ generated when electric current $I$ flows through a conductor with resistance $R$ for time $T$. Formally, a mathematical expression $\phi$ connects a set of input variables $\mathbf{x}$ and a set of constants $\mathbf{c}$ by mathematical operators. The possible mathematical operators include addition, subtraction, multiplication, division, trigonometric functions, etc. The meaning of these mathematical expressions follows their standard arithmetic definition.

Given a dataset ${D}=\{(\mathbf{x}_i, y_i):\mathbf{x}_i\in \mathbb{R}^n,y_i\in \mathbb{R}\}$ with $N$ samples, symbolic regression searches for the optimal expression $\phi^*$, such that $\phi^*(\mathbf{x}_i,\mathbf{c})\approx y_i$. From an optimization perspective, $\phi^*$ minimizes the averaged loss on the dataset:
\begin{equation} \label{eq:objective}
\phi^*\leftarrow\arg\min_{\phi\in \Pi}\;\frac{1}{N} \sum_{i=1}^{N} \ell(\phi(\mathbf{x}_i,\mathbf{c}), y_i),
\end{equation}
Here, hypothesis space $\Pi$ is the set of all candidate mathematical expressions; $\mathbf{c}$ denotes the constant coefficients in the expression; 
$\ell(\cdot,\cdot)$ denotes a loss function that penalizes the difference between the output of the candidate expression $\phi(\mathbf{x}_i,\mathbf{c})$ and the ground truth $y_i$. 
The set of all possible expressions i.e., the hypothesis space $\Pi$, can be exponentially large. As a result, finding the optimal expression is challenging and is shown to be NP-hard~\cite{journal/tmlr/virgolin2022}.

\textbf{Deep Policy Gradient for Symbolic Regression.} Recently, a line of work proposes the use of deep reinforcement learning (RL) for searching the governing equations~\cite{DBLP:conf/iclr/PetersenLMSKK21,DBLP:journals/corr/abs-1801-03526,DBLP:conf/nips/MundhenkLGSFP21}. They represent expressions as binary trees, where the interior nodes correspond to mathematical operators and leaf nodes correspond to variables or constants. The key idea is to model the search of different expressions, as a sequential decision-making process for the preorder traversal sequence of the expression trees using an RL algorithm. A reward function is defined to measure how well a predicted expression can fit the dataset. The deep recurrent neural network (RNN) is used as the RL learner for predicting the next possible node in the expression tree at every step of decision-making. The parameters of the RNN are trained using a policy gradient objective.

\textbf{Control Variable Experiment} studies the relationship between a few input variables and the output in the regression problem,  with the remaining input variables fixed to be the same~\cite{lehman2004designing}. 
In the controlled setting,  the ground-truth equation behaves the same after setting those controlled variables as constants, which is noted as the \textit{reduced-form equation}.
For example, the ground-truth equation  $\phi=x_1\times x_3-x_2\times x_4$ in Figure~\ref{fig:motivation}(a) is reduced to $x_1\times C_1-C_2$ when controlling $x_2,x_3,x_4$.  Figure~\ref{fig:motivation}(b,c) presents other reduced-form equations when the control variables are changed.
For the corresponding dataset ${D}$, the controlled variables are fixed to one value and the remaining variables are randomly assigned.  See Figure~\ref{fig:motivation}(a,b,c) for example datasets generated under different controlling variables. 

\textbf{Vertical Symbolic Regression} starts by finding a symbolic equation involving a small subset of the input variables and iteratively expands the discovered expression by introducing more variables. VSR relies on the control variable experiments introduced above.

VSR-GP was the first implementation of vertical symbolic regression using genetic programming (GP) ~\cite{DBLP:conf/pkdd/JiangX23,jiang2023vertical}. To fit an expression of $n$ variables, VSR-GP initially only allows variable $x_1$ to vary and controls the values of all the rest variables.
Using GP as a subroutine, VSR-GP finds a pool of expressions $\{\phi_{1}, \ldots, \phi_{m}\}$ which best fit the data from this controlled experiment. 
Notice 
$\{\phi_{1}, \ldots, \phi_{m}\}$ are restricted 
to contain only one free variable $x_1$ and $m$ is the pool size. A small error indicates $\phi_{i}$ is close to the ground truth reduced to the one free variable and thus is marked unmutable by the genetic operations in the following rounds.  In the 2nd round, VSR-GP adds a second free variable $x_2$ and starts fitting $\{\phi'_{1}, \ldots, \phi'_{m}\}$ using the data from control variable experiments involving the 
 two free variables $x_1,x_2$. After $n$ rounds,  the expressions
in the VSR-GP pool consider all $n$ variables.  
Overall, VSR expedites the discovery process because the first few rounds of VSR are significantly cheaper than the traditional \textit{horizontal} discovery process, which searches for optimal expression involving {all input variables at once}.

%% file: tex/3.pipeline.tex
\section{Methodology}
\noindent\textbf{Motivation.}
The prior work of VSR-GP uses genetic programming to edit the expression tree. GP is not allowed to edit internal nodes in the best-discovered expression trees from the previous vertical discovery step, to ensure later genetic operations do not delete the prior knowledge on the governing equation.  But, this idea cannot be easily integrated with deep reinforcement learning-based symbolic regressors. 

Employing deep neural nets in vertical symbolic regression to predict the symbolic equation tree in each vertical expansion step will result in (1) difficulty passing gradients from trees to deep neural nets. 
(2) complications in concatenating deep networks for predictions in each vertical expansion step. See two possible integrations in appendix~\ref{apx:direct-integ}. 

Our idea is to consider a new representation of expression. We extend the context-free grammar definition for symbolic expression, where a sequence of grammar rules uniquely corresponds to an expression.
We regard the prediction of symbolic expression as the sequential decision-making process of picking the sequence of grammar rules step-by-step. The RNN predicts grammar rules instead of nodes in the expression tree. 
The best-discovered reduced-form equation is converted into the start symbol in the grammar, ensuring the predicted expression from our RNN is always compatible with the prior knowledge of the governing equation.  This allows us to shrink the hypothesis space and accelerate scientific discovery because other non-reducible expressions will be never sampled from the RNN.

\noindent\textbf{Deep Vertical Symbolic Regression Pipeline.}
Figure~\ref{fig:hv} shows our deep vertical symbolic regression (\method) pipeline. The high-level idea of \method is to construct increasingly complex symbolic expressions involving an increasing number of independent variables based on control variable experiments with fewer and fewer controlled variables.  

To fit an expression of $n$ variables, we first hold all $n-1$ variables as constant and allow only one variable to vary.
We would like to find the best expression $\phi_{1}$, which best fits the data in this controlled experiment.  We use the deep RNN as the RL learner to search for the best possible expression. 
It is achieved by using the RNN to sample sequences of grammar rules defined for symbolic expression. Every sequence of rules is then converted into an expression, where the constants in the expression are fitted with the dataset. The parameters of the RNN model will be trained through the policy gradient objective. The expression with the best fitness score is returned as the prediction of the RNN. A visualized process is in Figure~\ref{fig:hv}(a, e).

Following the idea in VSR,
the next step is to decide whether each constant is a summary constant or a standalone constant. (1) A constant that is not relevant to any controlled variables is considered as standalone, which will be preserved in the rest rounds.  (2) A constant that is actually a sub-expression involving those controlled variables is noted as the summary type, which will be expanded in the following rounds. In our implementation, if the variance of fitted values of constant across multiple control variable experiments is high, then it is classified as the summary type. Otherwise, it is classified as a standalone type. 

Assuming we find the correct reduced-from equation $\phi_1$ after several learning epochs.
To ensure \method does not forget this discovered knowledge of the first round, we want all the expressions to be discovered in the following rounds can be reduced to $\phi_1$. Therefore, we construct $\phi_1$ as the start symbol for the following round by replacing every summary constant in $\phi_1$ as a non-terminal symbol in the grammar (noted as $A$), indicating a sub-expression.  For the example case in Figure~\ref{fig:motivation}(a), both of them are summary constants so the 1st round best-predicted expression $x_1\times C_1-C_2$ is converted as the 2nd round start symbol $x_1\times A-A$.

In the 2nd round, \method adds one more free variable and starts fitting $\phi_{2}$ using the data from control variable experiments involving two free variables. 
 Similar to the first round, we are restricted to only searching for sub-expressions with the second 
  variable. It is achieved by limiting the output vocabulary of the RNN model.
 In Figure~\ref{fig:motivation}(f), the RNN model finds an expression $\phi_2=x_1\times C_3-(x_2\times C_4)$. The semantic is that the RL learner learns to expand $C_1$ with another constant $C_3$ and $C_2$ with sub-expression $(x_2\times C_4)$, based on the best-discovered result of the 1st round $\phi_1=x_1\times C_1-C_2$.

Our \method introduces one free variable at a time and expands the equation learned in the previous round to include this newly added variable. 
This process continues until all the variables are considered. 
After $n$ rounds, we return the equations with the best fitness scores. The predicted equation will be evaluated on data with no variable controlled.
See the steps in Figure~\ref{fig:hv}(b,c,d) for a visual demonstration. We summarize the whole process of \method in Algorithm~\ref{alg:main} in the appendix.
The major difference of this approach from most state-of-the-art approaches is that those baselines learn to find the expressions in the full hypothesis space with all input variables, from a fixed dataset collected before training. Our \method accelerates the discovery process, because of the small size of the reduced hypothesis space, i.e., the set of candidate expressions involving only a few variables. The task is much easier than fitting the expression in the full hypothesis space involving all input variables.

%% file: tex/4.components.tex
\subsection{Expression Represented with Grammar Rules}
\label{sec:gram-rule}
We propose to represent symbolic expression by extending the context-free grammar (CFG)~\cite{DBLP:conf/icml/TodorovskiD97}. A context-free grammar is represented by a tuple $(V,\Sigma, R, S)$, where $V$ is a set of non-terminal symbols, $\Sigma$ is a set of terminal symbols, $R$ is a set of production rules and $S$ is a start symbol. 
In our formulation, (1) $\Sigma$ is the set of input variables and constants $\{x_1,\dots,x_n,\mathtt{const}\}$. (2) set of non-terminal symbols $V$ represents sub-expressions, like $\{A\}$. Here $A$ is a placeholder symbol.  (3) set of production rules $R$ represents mathematical operations such as addition, subtraction, multiplication, and division. That is $\{A\to (A+A),A\to (A-A),A\to A\times A,A\to A\div A\}$, where $\to$ represents the left-hand side 
 is replaced with its right-hand side.
(4) The start symbol $S$ is extended to be $A $, $x_1\times A-A$, or other symbols constructed from the best-predicted expression under the controlled variables.

\begin{figure}[!t]
    \centering
    \includegraphics[width=.59\linewidth]{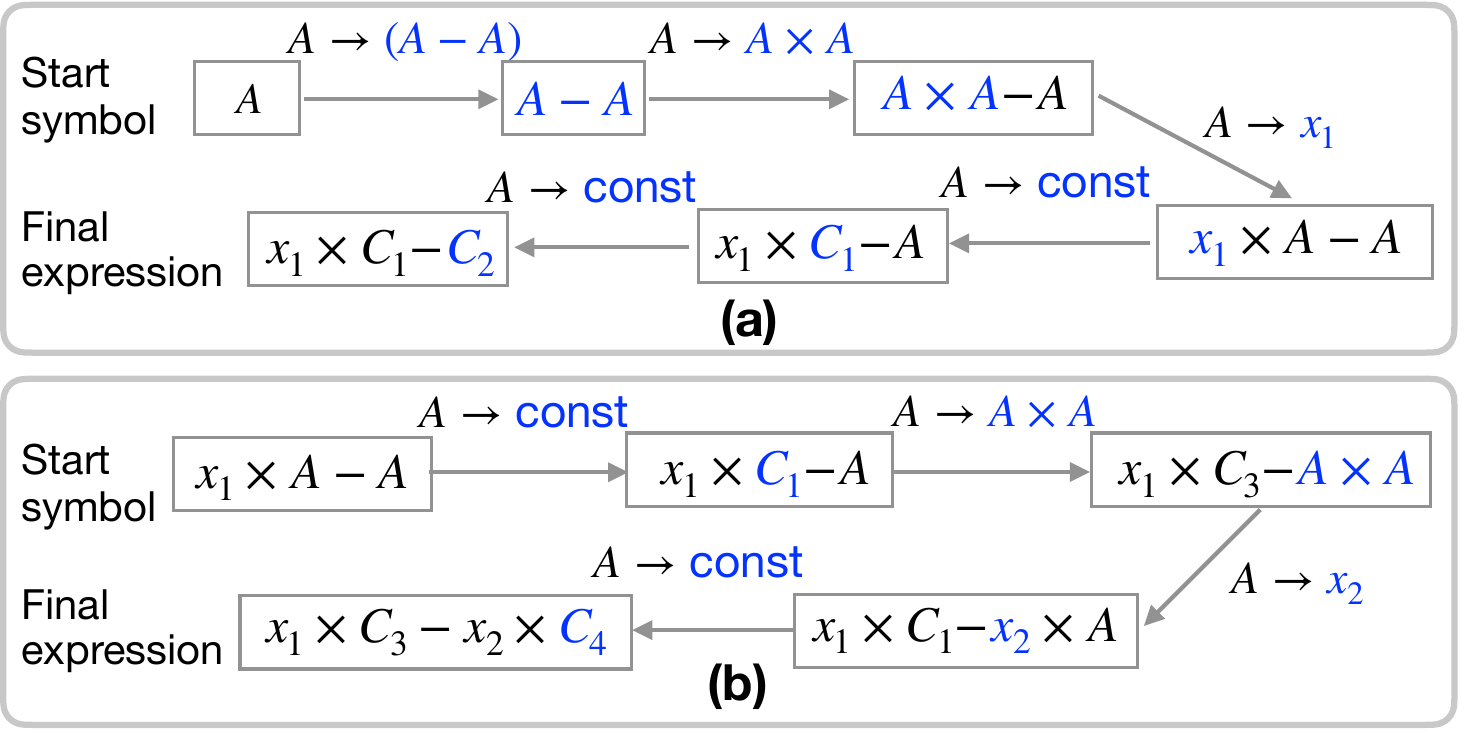}
    \caption{Convert a sequence of grammar rules into a valid expression. Each rule expands the first non-terminal symbol in the squared box. The parts that get expanded are color-highlighted.}
    \label{fig:expr-as-rules}
    \vspace{-.5em}
\end{figure}

Beginning with the start symbol, successive applying the grammar rules in different orders results in different expressions. Each rule expands the \textit{first} non-terminal symbol in the start symbol. An expression with only terminal symbols is a valid mathematical expression,  whereas an expression with a mixture of non-terminal and terminal symbols is an invalid expression.  
The expression can also be represented as a binary tree. We chose grammar representation instead of the binary tree for expressions because it will make the vertical symbolic regression process straightforward.

Figure~\ref{fig:expr-as-rules} presents two examples of constructing the expression $\phi$ from the start symbol using the given sequence of grammar rules. In Figure~\ref{fig:expr-as-rules}(a),
we first use the grammar rule for subtraction $A\to (A- A)$, which means that the symbol $A$ in expression $\phi=A$ is expanded with its right-hand side, resulting in $\phi = (A- A)$. By repeatedly using grammar rules to expand the non-terminal symbols, we finally arrive at our desired expression $\phi= x_1\times C_1-C_2$. In Figure~\ref{fig:expr-as-rules}(b), the start symbol is $x_1\times A-A$. The rule $A\to \texttt{const}$ replaces the first non-terminal symbol with a constant. Thus, we have $x_1\times C_1-A$. In the end, we obtain a valid expression $x_1\times C_3-x_2\times C_4$.

Afterward, we decide the optimal value of open constants in each expression. Assume the expression has $m$ open constants. We first sample a batch of data $D$ with the controlled variable $\mathbf{x}_c$ and then use a gradient-based optimizer to fit those open constants, by minimizing the objective $\min_{\mathbf{c}\in\mathbb{R}^m}\frac{1}{N}\sum_{i=1}^N\ell(\phi(\mathbf{x}_i,\mathbf{c}), y_i)$. We then obtain the fitness score $o$, the fitted constants $\mathbf{c}$, and the fitted equation $\phi$.

\subsection{Expression Sampling from Recurrent Network}~\label{sec:rnn-train}
\noindent\textbf{Vocabulary.}  In our vertical symbolic regression setting, the input and output vocabulary is the set of grammar rules that cover each input variable, constants, and mathematical operations.  We create an embedding layer for the input vocabulary, noted as the $\mathtt{Embd}$ function. For each input rule $r\in R$, its $d$-dimensional embedding vector is noted as $\mathbf{r}\in\mathbb{R}^d$.  

\noindent\textbf{Sampling Procedure.}
The RNN module samples an expression by sampling the sequence of grammar rules in a sequential decision-making process. Denote the sampled sequence of rules as $\tau=(\tau_1,\tau_2,\ldots)$. Initially, the RNN takes in the start symbol $\tau_1=S$ and computes the first step hidden state vector $\mathbf{h}_1$.
At $t$-th time step, RNN uses the predicted output from the previous step as the input of current step ${\tau}_t$. RNN computes its hidden state vector $\mathbf{h}_t$ using the embedding vector of input token $\tau_t$ and the previous time-step hidden state vector $\mathbf{h}_{t-1}$.  The linear layer and softmax function are applied to emit a categorical distribution $p(\tau_{t+1}|\tau_t,\mathbf{h}_{t-1})$ over every token in the output vocabulary, which represents the probability of the next possible rule in the half-completed expression $p(\tau_{t+1}|\tau_t,\ldots,\tau_1)$. The RNN samples one token from the categorical distribution $\tau_{t+1}\sim p(\tau_{t+1}|\tau_t,\mathbf{h}_{t-1})$ as the prediction of the next possible rule. To conclude, the computational pipeline at the $t$-th step is shown below:
\begin{equation} \label{eq:rnn-predict}
\begin{aligned}
\bm{\tau}_t &=\mathtt{Embd}(\tau_t), \\
\mathbf{h}_t &=\mathtt{RNN}(\bm{\tau}_t,\mathbf{h}_{t-1};\theta), \\
\mathbf{s}_t&=W\mathbf{h}_t+\mathbf{b}, \\
p(\tau_{t+1}\mbox{=}r_i|\bm{\tau}_t,\mathbf{h}_{t-1}) &=\frac{\exp( \mathbf{s}_{t,i})}{\sum_{r_j\in R}\exp(\mathbf{s}_{t,j})},\;\text{ for } r_i\in R. 
\end{aligned}
\end{equation}
The weight matrix $W\in\mathbb{R}^{|d|\times |R|}$ and bias vector $b\in\mathbb{R}^d$ are the parameters of the linear layer and the last row in Eq.~\ref{eq:rnn-predict} is the softmax layer. $\theta$ are the parameters of the RNN.
The sampled rule $r_{t+1}$ will be the input for the $t+1$-th step.

After $L$ steps, we obtain the sequence of rules $\tau=(\tau_1,\ldots, \tau_L)$ with probability $p(\tau|\theta)=\prod_{t=1}^{L-1}p(\tau_{t+1}|\tau_1,\ldots, \tau_{t};\theta)$. We convert this sequence into an expression by following the procedure described in Section~\ref{sec:gram-rule}.  For cases where we arrive at the end of the sequence while there are still non-terminal symbols in the converted expression, we would randomly add some rules with only terminal symbols to complete the expression. For cases where we already get a valid expression in the middle of the sequence, we ignore the rest of the sequence and return the valid expression.


\noindent\textbf{Policy Gradient-based Training.} 
We follow the reinforcement learning formulation to train the parameters of the RNN module~\cite{DBLP:journals/igpl/WierstraFPS10}.  The sampled rules before the current step $t$, i.e., ($\tau_1,\ldots,\tau_t)$, is viewed as the \textit{state} of the $t$-th step for the RL learner. Those rules in the output vocabulary are the available \textit{actions} for the RL learner. In the formulated decision-making process, the RNN takes in the current state and outputs a distribution over next-step possible actions. The objective of the RL learner is to learn to pick the optimal sequences of grammar rules to maximize the expected rewards.
Denote the converted expression from  $\tau$ as $\phi$. A typical reward function is defined from the fitness score of the expression $\mathtt{reward}(\tau)=1/(1+\mathtt{NMSE}(\phi))$. The objective that maximizes the expected reward from the RNN model is defined as $J(\theta)=\mathbb{E}_{\tau\sim p(\tau|\theta)}(\mathtt{reward}(\tau))$, where $p(\tau|\theta)$ is the probability of sampling sequence $\tau$ from the RNN. 

The gradient with respect to the objective $\nabla_{\theta}J(\theta)$ needs to be estimated. 
We follow the classic REINFORCE policy gradient algorithm~\cite{DBLP:journals/ml/Williams92}. We first sample several times from the RNN module and obtain $N$ sequences $(\tau^1,\ldots, \tau^N)$, an unbiased estimation of the gradient of the objective is computed as $\nabla_{\theta}J(\theta)\approx \frac{1}{N}\sum_{i=1}^N\mathtt{reward}(\tau^i)\nabla_{\theta}\log p(\tau^i|\theta)$. The parameters of the deep network are updated by the gradient descent algorithm with the estimated policy gradient value. In the literature, several practical tricks are proposed to reduce the estimation variance of the policy gradient. A common choice is to subtract a baseline function $b$ from the reward, as long as the baseline is not a function of the sample batch of expressions. Our implementation adopts this trick and the detailed derivation is presented in Appendix~\ref{apx:vsr}.
There are many variants like risk-seeking policy gradient~\cite{DBLP:conf/iclr/PetersenLMSKK21}, priority queue training~\cite{DBLP:journals/corr/abs-1801-03526}.

Throughout the whole training process, the expression with optimal fitness score from all the sampled expressions is used as the prediction of \method at the current round.

\subsection{Construct Start Symbol from the best-predicted Expression} \label{sec:distill}
Given the best-predicted expression $\phi$ and controlled variables $\mathbf{x}_c$, the following step is to construct the start symbol of the next rounds. This operation ensures all the future expressions can be reduced to any previously discovered equation thus all the discovered knowledge is remembered. It expedites the discovery of symbolic expression since other expressions that cannot be reduced to $\phi$ will be never sampled from the RNN. It requires first classifying the type of every constant in the expression into stand-alone or summary type, through multi-trail control variable experiments. Then we replace each summary constant with a placeholder symbol (i.e., ``$A$'') indicating a sub-expression containing controlled variables.

Following the procedure proposed in~\cite{DBLP:conf/pkdd/JiangX23}, we first query $K$ data batches  $(D_1,\ldots,D_K)$ with the same controlled variables  $\mathbf{x}_c$.  The controlled variables take the same value within each batch while taking different values across data batches.
We fit open constants in the candidate expression $\phi$ with each data batch by the gradient-based optimizer, like BFGS~\cite{fletcher2000practical}. We obtain multiple fitness scores $(o_1,\ldots, o_K)$ and multiple solutions to open constants $(\mathbf{c}_1,\ldots,\mathbf{c}_K)$. By examining the outcomes of $K$-trials control variable experiments, we have:
(1) Consistent close-to-zero fitness scores imply the fitted expression is close to the ground-truth equation in the reduced form. 
That is $o_k\le \varepsilon$ for all $1\le k\le K$, where  $\varepsilon$ is the threshold for the fitness scores.
(2) Conditioning on the result in case (1),  the $j$-th open constant is a standalone constant when the empirical variance of its fitted values across $K$ trials is less than a threshold $\varepsilon'$.  
In practice, if the best-predicted expression by the RNN module is not consistently close to zero, then all the constants in the expression are summary constants. Finally, the start symbol is obtained by replacing every summary constant with the symbol ``$A$''  according to our grammar.

%% file: tex/5.related.tex
\input{tex/6.1.algebraic}

\section{Related Work}
Recently AI has been highlighted to enable scientific discoveries in diverse domains~\cite{doi:10.1126/science.abj6511,jumper2021highly,wang2023scientific}. 
Early work in this domain focuses on learning logic (symbolic) representations~\cite{BRADLEY2001reasoning}.
Recently, there has been extensive research on learning algebraic equations~\cite{DBLP:conf/iclr/PetersenLMSKK21,DBLP:conf/nips/MundhenkLGSFP21} and differential learning differential equations from data~\cite{Dzeroski1995lagrange,brunton2016sparse,PhysRevE.100.033311,doi:10.1098/rspa.2018.0305,iten2020discovering,DBLP:conf/nips/CranmerSBXCSH20,Raissi20Fluid,RAISSI2019PhysicsInformedNN,Liu21AIPoincare,nanovoid_tracking,chen2018neural}. 
In this domain, a line of works develops robots that automatically refine the hypothesis space, some with human interactions~\cite{langey1988scientificdiscovery,Valdes1994,king2004functional,king2009autosci}.
These works are quite related to ours because they also actively probe the hypothesis spaces, albeit they are in biology and chemistry.


Existing works on multi-variable regression are mainly based on pre-trained encoder-decoder methods with massive training datasets (e.g., {millions of data points}~\cite{DBLP:conf/icml/BiggioBNLP21}), and even larger-scale generative models (e.g., approximately 100 million parameters~\cite{DBLP:conf/nips/KamiennydLC22}).
Our \method algorithm is a tailored algorithm to solve multi-variable symbolic regression problems. 

The idea of using a control variable experiment tightly connects to the BACON system~\cite{DBLP:conf/ijcai/Langley77,DBLP:conf/ijcai/Langley79,DBLP:conf/ijcai/LangleyBS81,king2004functional,king2009autosci,cerrato2023rlsci}. Our method develops on the current popular deep recurrent neural network while their method is a rule-based system, due to the historical limitation.

Our method connects to the symbolic regression method using probabilistic context-free grammar~\cite{DBLP:conf/icml/TodorovskiD97,DBLP:journals/kbs/BrenceTD21,DBLP:conf/dis/GecOBDT22}. They use a fixed probability to sample rules, we use a deep neural network to learn the probability distribution.

Our \method is also tightly connected to deep symbolic regression~\cite{DBLP:conf/iclr/PetersenLMSKK21,DBLP:conf/nips/MundhenkLGSFP21}. We both use deep recurrent networks to predict a sequence of tokens that can be composed into a symbolic expression. However, their method predicts the preorder traversal sequence for the expression tree while our method predicts the sequence of production rules for the expression.


%% file: tex/6.1.algebraic.tex
\begin{table*}[!t]
    \centering
    \label{tab:Trigonometric-nmse-noiseless}
     \begin{tabular}{r|cccccccc}
    \hline

Methods & $(2, 1, 1)$ & $(3, 2, 2)$ & $(4, 4, 6)$ & $(5,5,5)$ & $(5,5,8)$ & $(6,6,8)$ & $(6,6,10)$ & $(8, 8, 12)$\\   \midrule
\cvgp & $0.005$ & $0.028$ & $0.086$ & $0.014$ & $0.066$ & $0.066$ & $\mathbf{0.104}$  & T.O. \\
GP & $7{E-}4$ & $0.023$ & $0.044$ & $0.063$ & $0.102$ & $0.127$ & $0.159$ & 0.872\\
Eureqa & $<$\textbf{1E-6} & $<$\textbf{1E-6} & $0.024$ & $0.158$ & $0.284$ & $0.433$ & $0.910$ & $0.162$ \\ 
\hline
SPL & $0.006$ & $0.033$ & $0.144$ & $0.147$ & $0.307$ & $0.391$ & $0.472$ & $0.599$\\
E2ETransformer & 0.018 & 0.0015 & 0.030 & 0.121 & $0.072$ & $0.194$ &  0.142 &   0.112 \\ 
\hline
DSR & $<\textbf{1E-6}$ & $0.008$ & $2.815$ & $2.558$ & $2.535$ & $0.936$ & $6.121$ & $0.335$ \\
PQT & $0.020$ & $0.161$ & $2.381$ & $2.168$ & $2.482$ & $0.983$ & $5.750 $ & $0.232$\\
VPG & $0.030$ & $0.277$ & $2.990$ & $1.903$ & $2.440$ & $0.900$ & $3.857$ & $0.451$\\
GPMeld & $<1{E-}6$ & $0.112$ & $1.670$ & $1.501$ & $2.422$ & $0.964$ & $7.393$ & T.O.\\
\method (ours) & $<\textbf{1E-6}$  & $<\textbf{1E-6}$  & $<\textbf{1E-6}$ &  $<\textbf{1E-6}$ & $\mathbf{0.026}$ & $\mathbf{0.063}$ & 0.114 & $\mathbf{0.101}$\\
\hline
    \end{tabular}
    \caption{On selected algebraic equation datasets, median (50\%-quartile) of NMSE values of the best-predicted expressions found by all the algorithms. The set of mathematical operator is $O_p=\{+,-,\times,\sin,\cos,\texttt{const}\}$. The 3-tuples at the top $(\cdot,\cdot,\cdot)$ indicate the number of free variables, singular terms, and cross terms in the ground-truth expressions generating the dataset. $O_p$ stands for the set of allowed operators.  ``T.O.'' implies the algorithm is timed out for
48 hours.} \label{tab:trig}
\vspace{-1.2em}
\end{table*}

%% file: tex/6.exp.tex
\section{Experiment}
In this section, we evaluate the performance of the proposed \method method on several multi-variable algebraic equations datasets and further extend to real-world differential equation discovery tasks. 

\subsection{Symbolic Regression on Algebraic Equations}
\noindent\textbf{Experiment Settings.} For the dataset on algebraic expressions, we consider the  8 groups of expressions from the Trigonometric dataset~\cite{DBLP:conf/pkdd/JiangX23}, where each group contains 10 randomly sampled expressions.  In terms of baselines, 
 we consider (1) evolutionary algorithm:  Genetic Programming (GP),  Control Variable Genetic Programming (CVGP)~\cite{DBLP:conf/pkdd/JiangX23}, and Eureqa~\cite{DBLP:journals/gpem/Dubcakova11}.
(2) deep reinforcement learning:  Priority queue training (PQT)~\cite{DBLP:journals/corr/abs-1801-03526}, Vanilla Policy Gradient (VPG)~\cite{DBLP:journals/ml/Williams92}, Deep Symbolic Regression (DSR)~\cite{DBLP:conf/iclr/PetersenLMSKK21}, and Neural-Guided Genetic Programming Population Seeding (GPMeld)~\cite{DBLP:conf/nips/MundhenkLGSFP21}. (3) Monte Carlo Tree Search: Symbolic Physics Learner (SPL)~\cite{DBLP:conf/iclr/Sun0W023}, (4) Transformer network with pre-training: end-to-end Transformer (E2ETransformer)~\cite{DBLP:conf/nips/KamiennydLC22}. In terms of evaluation metrics, we evaluate the normalized-mean-squared-error (NMSE) of the best-predicted expression by each algorithm, on a separately-generated testing dataset $D_{\text{test}}$.  We report median values instead of means due to outliers. Symbolic regression belongs to combinatorial optimization problems, which commonly have no mean values. The detailed experiment configurations are in Appendix~\ref{apx:alge-eq}.

\noindent\textbf{Goodness-of-fit Comparison.} We consider our \method against several challenging datasets involving multiple variables. 
In table~\ref{tab:trig}, We report the median NMSE on the selected algebraic datasets. Our \method attains the smallest median NMSE values in 7 out of 8 datasets, against a line of current popular baselines including the original VSR-GP. The main reason is deep networks offer many more parameters than the GP algorithm, which can better adapt to different datasets and sample higher-quality expressions from the deep networks.

\noindent\textbf{Extended Large-scale Comparison.} In the real world, scientists may collect all available variables that are more than needed into symbolic regression, where only part of the inputs will be included in the ground-truth expression. 
We randomly pick $5$ variables from all the $n$ variables and replace the appeared variable in expressions confined as $(5,5,5)$ in Table~\ref{tab:trig}. In Table~\ref{tab:extend-trig}, we collect the median NMSE values on this large-scale dataset setting. Our \method scales well because it first detects all the contributing inputs using the control variable experiments. Notice that those baselines that are easily timeout in this setting are excluded for comparison.

\begin{table}[!t]
    \centering
    \small
    \begin{tabular}{r|c|c|c|c|c}
    \hline
    &\multicolumn{5}{c}{Total Input Variables} \\
           & $n=10$ & $n=20$ &  $n=30$ &  $n=40$ & $n=50$ \\ \hline
           
           SPL & 0.386 & 0.554&0.554 & 0.714& 0.815\\
           GP & 0.159 & 0.172 & 0.218 & 0.229 & 0.517\\
           DSR & 0.284	& 0.521	& 0.522	& 0.660	& 0.719\\
           VPG & 0.415	& 0.695	& 0.726	& 0.726	& 0.779\\
           PQT & 0.384	& 0.488	& 0.615	& 0.620	& 0.594\\
           \hline
     \method   &  $<\textbf{1E-6}$ &  $<\textbf{1E-6}$  & $<\textbf{1E-6}$  &\textbf{0.002}&  \textbf{0.021}  \\
     \hline
    \end{tabular}
      \caption{Median NMSE values on extended large-scale algebraic equation dataset. Our \method scales better to more variable settings than baselines due to the control variable experiment. $n$ is the total variables in the dataset.}
    \label{tab:extend-trig}
    \vspace{-1.4em}
\end{table}

\noindent\textbf{Exact Recovery Comparison.} We compare if each learning algorithm finds the exact equation, the result of which is collected in Table~\ref{tab:recovery-rate}. The discovered equation by each algorithm is further collected in Appendix~\ref{apx:exact-result}.  We can observe that our \method has a higher rate of recovering the ground-truth expressions compared to baselines. This is because our method first use a control variable experiment to pick what are the contributing variables to the data and what are not.
\begin{table}[!ht]
    \centering
    \begin{tabular}{c|cccc}
    \hline
     & $(2, 1, 1)$ & $(3, 2, 2)$ & $(4, 4, 6)$ & $(5,5,5)$ \\\hline
SPL & 	$20\%$	& $10\%$	& $0\%$	& $0\%$\\
\footnotesize{E2ETransformer} & $0\%$	& $0\%$ & 	$0\%$ & 	$0\%$\\
VSR-GP	& $60\%$	& $50\%$ & 	$0\%$	& $0\%$ \\
VSR-DPG	& $\mathbf{100\%}$	& $\textbf{70\%}$	& $\textbf{60\%}$	& $\mathbf{40\%}$ \\
\hline
    \end{tabular}
    \caption{On selected algebraic equations, the exact recovery rate over the best-predicted found by all the algorithms. Our \method has a higher rate of recovering the ground-truth expressions compared to baselines.}
    \label{tab:recovery-rate}
    \vspace{-1.4em}
\end{table}

\subsection{Symbolic Regression on Differential Equations} 
\noindent\textbf{Task Definition.} The temporal evolution of the dynamic system is modeled by the time derivatives of the state variables. Let $\mathbf{x}$ be the $n$-dimensional vector of state variables, and $\dot{\mathbf{x}}$ is the vector of their time derivatives. The differential equation is of the form $\dot{\mathbf{x}}=\phi(\mathbf{x},\mathbf{c})$, where constant vector $\mathbf{c}\in\mathbb{R}^m$ are parameters of the dynamic system. Following the definition of symbolic regression on differential equation in~\cite{DBLP:conf/dis/GecOBDT22,DBLP:conf/iclr/Sun0W023},
given a trajectory dataset of state variable and its time derivatives $\{(\mathbf{x}(t_i),\dot{\mathbf{x}}(t_i))\}_{i=1}^N$,  the symbolic regression task is to predict the best expression $\phi(\mathbf{x},\mathbf{c})$ that minimizes the average loss on trajectory data: $\arg\min_{\phi}\frac{1}{N}\sum_{i=1}^N\ell(\dot{\mathbf{x}}(t_i),\phi(\mathbf{x}(t_i),\mathbf{c}))$. Other formulation of this problem assume we have no access to its time derivatives, that is $\{(t_i, \mathbf{x}(t_i))\}_{i=1}^N$~\cite{2023odeformer}.

\noindent\textbf{Experiment Setting.} We consider recent popular baselines for differential equations, including (1) SINDy~\cite{brunton2016sparse}, (2) ODEFormer~\cite{2023odeformer}, (3)  Symbolic Physics Learner (SPL)~\cite{DBLP:conf/iclr/Sun0W023}. 4) Probabilistic grammar for equation discovery (ProGED)~\cite{DBLP:journals/kbs/BrenceTD21}.
In terms of the dataset, we consider the Lorenz Attractor with $n=3$ variables, Magnetohydrodynamic (MHD) turbulence with $n=6$ variables, and Glycolysis Oscillation with $n=7$ variables. All of them are collected from~\cite{brunton2016sparse}.
To evaluate whether the algorithm identifies the ground-truth expression, we use the Accuracy metric based on the coefficient of determination ($R^2$).  The detailed experiment configurations are in Appendix~\ref{apx:diff-eq}.

\begin{figure}[!t]
    \centering
    \includegraphics[width=.34\linewidth]{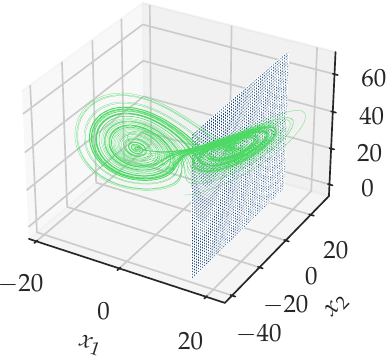}
    \includegraphics[width=.34\linewidth]{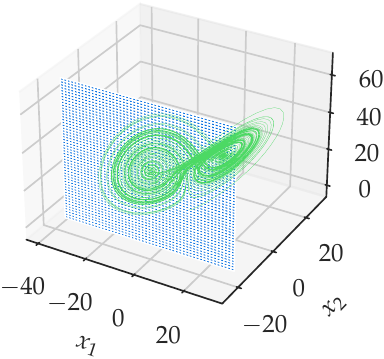}
    \caption{Visualization of \method controlling variables $x_1$ (Left) and $x_2$ (Right) for the Lorenz attractor. The data of our \method are drawn from the intersection of the mesh plane and the curve on the Lorenz attractor. In comparison, the ODEFormer draws data by picking a consecutive sequence $\{(t_i,\mathbf{x}(t_i))\}_{t=0}^N$ without knowing its time derivative on the curve.}
    \label{fig:lorenz-control}
    \vspace{-1em}
\end{figure}

\noindent\textbf{Result Analysis.} The results are summarized in Table~\ref{tab:diff-equation}. Our proposed \method discovers a set of differential expressions with much higher quality than the considered baselines. We further provide a visual understanding of the proposed \method method in Figure~\ref{fig:lorenz-control}. The data of our \method are drawn from the intersection of the mesh plane and the curve on the Lorenz attractor. In comparison, the current baselines draw data by picking a random trajectory or many random points on the curve. We notice the ODEFormer is pre-trained on differential equations up to two variables, thus does not scale well with more variable settings. The predicted differential equations by each algorithm are in appendix~\ref{apx:tab:diff-eq-exact}.

 \begin{table}[!t]
     \centering
     \begin{tabular}{r|c|c|c}
    \hline
         & Lorenz&  MHD & Glycolysis    \\ \hline
             SPL& $\mathbf{100\%}$ &   50\%  & $14.2\%$  \\
         SINDy & $\mathbf{100\%}$ & 0\%  & $0\%$\\
         ProGED &  $0\%$ & $0\%$ & 0\% \\
            ODEFormer &  0\%  & 0\% & NA \\
         \method(ours)&  $\mathbf{100\%}$   & $\mathbf{100\%}$  &  $\mathbf{87\%}$ \\
    \hline
    \end{tabular}
    \caption{On the differential equation dataset, ($R^2\ge 0.9999$)-based accuracy is reported over the best-predicted expression found by all the algorithms. Our \method method can discover the governing expressions with a much higher accuracy rate than baselines.}
     \label{tab:diff-equation}
     \vspace{-1em}
 \end{table}

%% file: tex/7.conclude.tex
\section{Conclusion}
In this research, we propose Vertical Symbolic Regression with Deep Policy Gradient (i.e., \method) to discover governing equations involving many independent variables, which is beyond the capabilities of current state-of-the-art approaches.
\method follows a vertical discovery path -- it builds equations involving more and more input variables using control variable experiments. Because the first few steps following the vertical discovery route are much cheaper than discovering the equation in the full hypothesis space, \method has the potential to supercharge current popular approaches. Experimental results show \method can uncover complex scientific equations with more contributing factors than what current approaches can handle.

%% file: tex/10.impelment.tex
\section{Direct Integration of Vertical Symbolic Regression \\ with Deep Policy Gradient} \label{apx:direct-integ}

Here we provide several possible pipelines for integrating the idea of vertical symbolic regression with deep reinforcement learning, using the binary tree representation of symbolic expressions. We will show the limitations of each integration. The fundamental cause is the tree representation of expression.

\begin{figure*}[!t]
    \centering
    \includegraphics[width=0.95\linewidth]{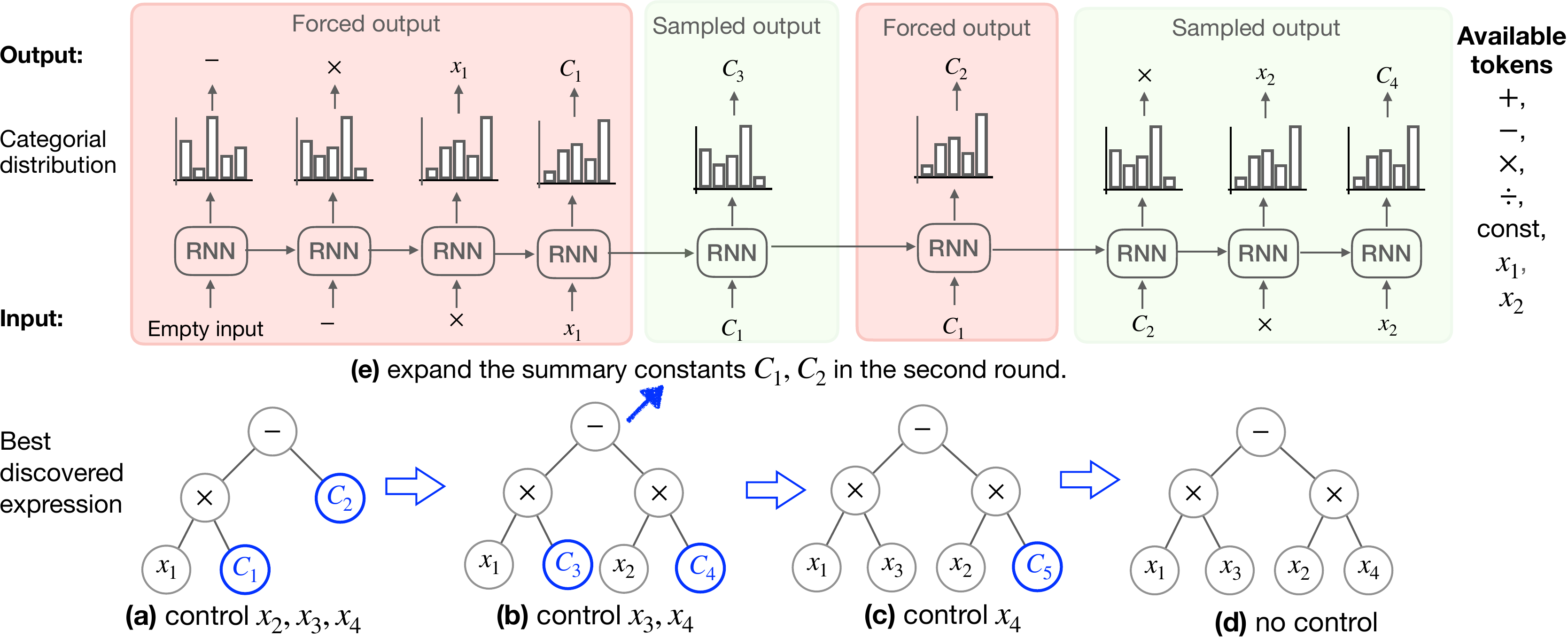}
    \caption{Constraint-based integration of deep reinforcement learning with vertical symbolic regression. The constraints enforce the output of RNN output the given token at each step. It has limitations in passing the gradient to the parameters of RNN and also requires heavy engineering of different constraints. \textbf{(a)} Initially, the RNN to learn a reduced form equation with variables $x_2,x_3,x_4$ controlled. The RNN learns to sample the best preorder traversal of the reduced form expression tree from the \textit{available tokens}. No constraints are applied in the first round. \textbf{(b, e)} Given the best-predicted expression $\phi_1$ represented as  $(-,\times, x_1, C_1, C_2)$ at the first round, the RNN is used to predict an expression with control variables $x_3,x_4$. For the first four steps, the constraints are applied to mask out other tokens in the output, to enforce that the output must be $-,\times, x_1,C_1$. Since $C_1$ is a summary constant, the RNN samples a sub-expression with no constraints starting at the 5th step, which is $C_3$. In 6-th step, with the termination of the prior sub-expression, constraints are applied to enforce the RNN outputs $C_2$. Starting at the 7th step, we sample a subexpression $x_2\times C_4$.   \textbf{(c,d)} The rest steps in the pipeline of vertical symbolic regression using expression tree representation.}
    \label{fig:constraints}
\end{figure*}

\paragraph{Symbolic Expression as Tree}
A symbolic expression can be represented as an \textit{expression tree}, where variables and constants correspond to leaves, and operators correspond to the inner nodes of the tree. 
An inner node can have one or multiple child nodes depending on the arity of the associated operator. For example, a node representing the addition operation ($+$) has 2 children, whereas a node representing trigonometric functions like $\cos$ operation has a single child node. 
The preorder traversal sequence of the expression tree uniquely determines a symbolic expression. Figure~\ref{fig:constraints}(a) presents an example of such an expression tree of the expression $x_1\times C_1-C_2$. Its preorder traversal sequence is $(-,\times, x_1,C_1,C_2)$. This traversal sequence uniquely determines a symbolic expression.

\paragraph{Genetic Programming for Symbolic Regression} Genetic Programming (GP)~\cite{koza1994genetic} has been a popular algorithm for symbolic regression. 
The core idea of GP is to maintain a pool of expressions represented as \textit{expression trees}, and iteratively improve this pool according to the fitness score.
The fitness score of a candidate expression measures how well the expression fits a given dataset. 
Each generation of GP consists of 3 basic operations -- \textit{selection, mutation} and \textit{crossover}. In the \textit{selection} step, candidate expressions with the highest fitness scores are retained in the pool, while those with the lowest fitness scores are discarded. 
In the \textit{mutation} step, sub-expressions of some randomly selected candidate expressions are altered with some probability. In the \textit{crossover} step, the sub-expressions of different candidate expressions are interchanged with some probability.    In implementation, \textit{mutation} changes a node of the expression tree while \textit{crossover} is the exchange of subtrees between a pair of trees. 
This whole process repeats until we reach the final generation.
We obtain a pool of expressions with high fitness scores, i.e., expressions that fit the data well, as our final solutions. 

\paragraph{Genetic Programming for Vertical Symbolic Regression (\cvgp).}
\cvgp uses GP as a sub-routine to predict the best expression at every round. At the end of every round,
for an expression in the pool with close-to-zero MSE metric, \cvgp marks the inner nodes for mathematical operators and leaf nodes for standalone constants as non-mutable. Only the leaf nodes for summary constants as marked as mutable, because summary constants are those sub-expressions containing controlled variables. During \textit{mutation} and \textit{crossover}, \cvgp only alters the mutable nodes of the candidate expression trees. In classic GP, all the tree nodes are mutable.

\paragraph{Deep Policy Gradient for Symbolic Regression}
The deep reinforcement learning-based approaches predict the expression by sampling the pre-order traversal sequence of the expression using RNN. The parameters of the RNN are trained through a policy gradient-based objective. The original work~\cite{DBLP:conf/iclr/PetersenLMSKK21} proposes a relatively complex RL-based symbolic regression framework, where those extra modules are omitted in this part to ensure the main idea is clearly delivered.

In the following, we present three possible integration plans for the vertical discovery path with reinforcement learning-based symbolic regression algorithms.

\subsection{Constraint-based Integration}
The first idea is to apply constraints to limit the output vocabulary to force the predicted expression at the current round to be close to the previously predicted expression and also attain a close-to-zero expression given the controlled variables.

\begin{figure*}[!t]
    \centering
    \includegraphics[width=0.95\linewidth]{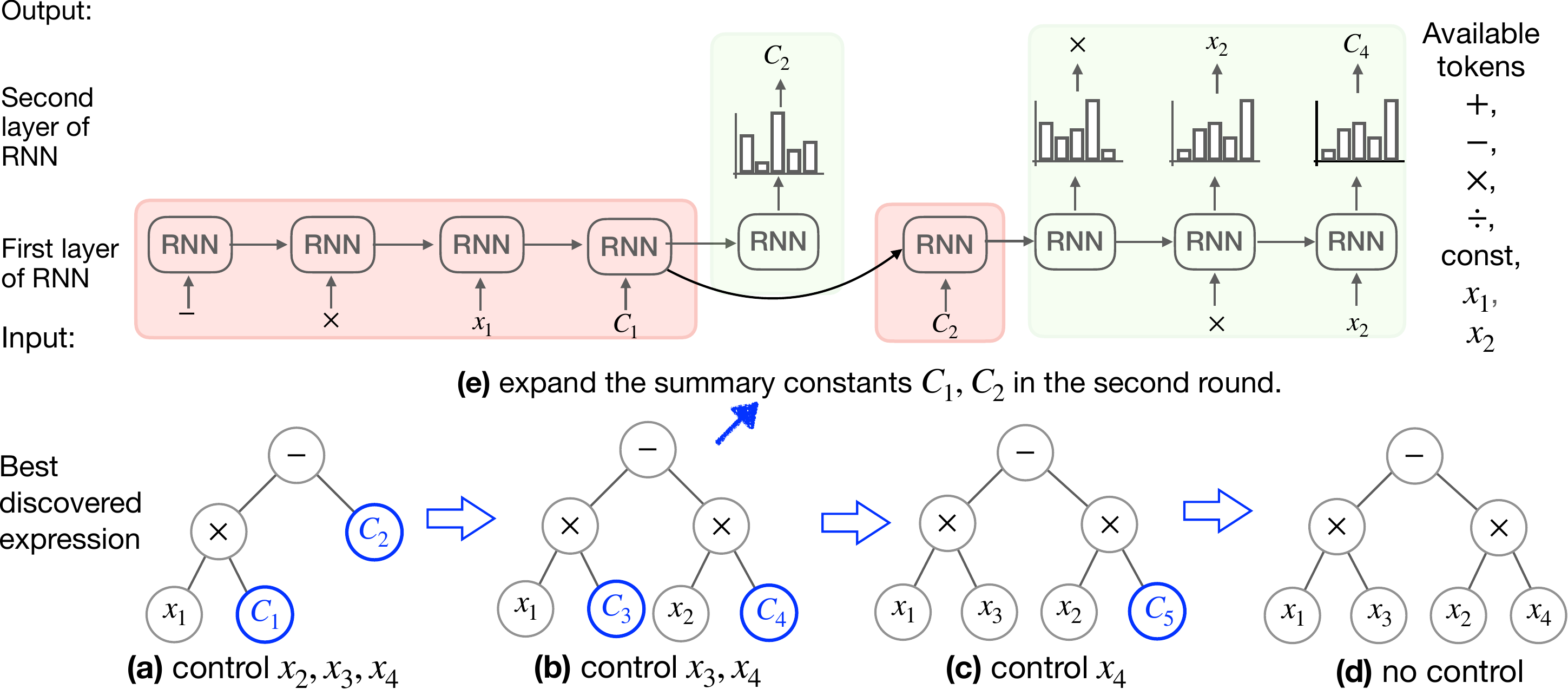}
    \caption{Concatenation-based integration of deep reinforcement learning with vertical symbolic regression. Multiple layers of RNN are concatenated together to implement the vertical symbolic regression. The limitation is we need to store all the parameters of previously trained RNN, leading to a very complicated approach.  \textbf{(a,b,c,d)} The pipeline of vertical symbolic regression using expression tree representation.  \textbf{(e)} The first layer takes the input of the best-predicted expression $\phi$, and the second layer uses the hidden vectors of the 4-th step and 5-th step of the first layer, as input to predict two separated sequence $C_3$ and $\times,x_2, C_4$. The parameters of the first layer are frozen while the parameters of the second layer are trained. }
    \label{fig:concat}
\end{figure*}

Take Figure~\ref{fig:constraints} as an example. Given the best-predicted expression $\phi_1$ represented as  $(-,\times, x_1,C_1,C_2)$ at the first round, we want to use the RNN to predict an expression $\phi$ that (1) has close-to-zero MSE value on the data with control variables $x_3,x_4$, and (2) similar to $\phi_1$ under controlled variables $x_2, x_3,x_4$. It could be achieved by forcing the RNN to predict $-$ at the first step, where the rest of the available tokens are masked out by the designed constant. 
Similarly, we force the RNN to predict the rest of two tokens $\times, \mathtt{const}$ with the designed constraints. Since we know the 4-th step output is a summary constant,  so we sample a token from the probability distribution. In the 6th step, the constraint is applied to force the RNN to output $C_2$, because the previous sub-expression has been completed. This constraint-based approach will force the sampled expression, like $\phi_2=x_1\times C_3-x_2\times x_4$, to be close to the best expression $\phi_{1}$ of the prior round.

The limitations are: (1) heavy engineering of designing the constraints and checking if the sub-expression has been completed. In Figure.~\ref{fig:constraints}(e), every step of constraints is different from the others. (2) Gradient computation issue. Only when the first sub-expression is done can we then apply constraints to enforce the RNN to output $C_2$, this will cause the gradient computation of the loss function to the parameters of RNN.

\subsection{Concatenation-based Integration}
The second possible idea is concatenating multiple layers of RNNs. The first layer of RNN is the trained RNN at the 1st round.  We use the first layer of the RNN to take in the best sequence of the first round. When we read in a summary constant, we use the updated hidden vector of the first layer as the initial vector of the second layer RNN.

Take Figure~\ref{fig:concat} as an example. The first layer RNN takes the sequence $(-,\times,\mathtt{const},x_1,\mathtt{const})$. Because the 4th and 4th step input is summary constant type,  we use the 4th and 5th step output vectors to initialize the hidden state vector of the second layer RNN. The second layer of RNN predicts two separated sub-expressions: $\mathtt{const}$ and $\times, x_2, \mathtt{const}$.  The whole sequence corresponds to the sampled expression $\phi_2$ from the concatenated RNNs.  The parameters inside the second layer RNN need to be trained by the policy gradient algorithm. Similarly, at the last round, we re-use pre-trained $n-1$ layers of RNN to take in the best-predicted expression $\phi_{n-1}$ and use one more layer to expand the summary constants in expression $\phi_{n-1}$. The whole predicted sequence of tokens is the final predicted expression $\phi_n$. 
Notice that the parameters of the prior layers of RNN can be frozen to reduce the number of parameters for training.

The main limitation of this idea is that: (1) we need to store all the trained RNNs. This does not scale up to many input variable cases. At the last round, we will have $n-1$ frozen layer of RNN and one trainable layer of RNN. (2) Due to multiple layers of RNNs and the sequence of input becoming longer and longer, then the training speed of the whole model will be slower and slower with fewer and fewer controlled variables.

\section{Extended Explanation of \method method} \label{apx:vsr}
\paragraph{Data-availability Assumption} 
A crucial assumption behind the success of vertical symbolic regression is the availability of a  $\mathtt{DataOracle}$ that returns a (noisy) observation of the dependent output with input variables in $\mathbf{x}_c$ controlled. 
Such a data oracle represents conducting control variable experiments in the real world, which can be expensive. 
This differs from the horizontal symbolic regression, where a dataset is obtained prior to learning \cite{doi:10.1080/15598608.2007.10411855} with no variable controlled.

The vertical discovery path is to build algorithms that mimic human scientific discovery, which has achieved tremendous success in early works~\cite{DBLP:conf/ijcai/Langley77,DBLP:conf/ijcai/Langley79,DBLP:conf/ijcai/LangleyBS81}. Recent work~\cite{chen2022generalisation,keren2023computational,DBLP:conf/gecco/HautBP22,DBLP:conf/gecco/HautPB23} also pointed out the importance of having a data oracle that can actively query data points, rather than learning from a fixed dataset.
In cases where it is difficult to obtain such a data oracle, \citeauthor{keren2023computational} proposed the use of deep neural networks to learn a data generator for the given set of controlled variables.

\begin{algorithm*}
   \caption{Vertical Symbolic Regression via Deep Policy Gradient.}\label{alg:main}  
   \begin{algorithmic}[1]
   \Require{ \#input variables $n$; Mathematical Operators $O_p$; Draw data with controlled variables $\mathtt{DataOracle}$.}
   \Ensure{The best-predicted expression.}
   \State $\mathbf{x}_c \gets \{x_1, \ldots, x_n\}$.\Comment{controlled variables}
   \State $S= A.$ \Comment{start symbol}
   \State $\mathcal{Q}\gets \emptyset.$ \Comment{best expressions across all rounds}
   \State ${{D}_{global}} \gets \mathtt{DataOracle}( \emptyset)$. \Comment{data oracle with no control variable}
   \State draw a batch of data $T_g\gets\mathtt{GenData}(D_{global})$. %
   \For{$x_i \in \{x_1, \ldots, x_n\}$ }
        \State set controlled variables $\mathbf{x}_c \gets \mathbf{x}_c \setminus \{x_i\}$. 
        \State construct data oracle $D_o \gets \mathtt{DataOracle}(\mathbf{x}_c)$. 
        \State $\phi \gets \textsc{Dpg}(S,D_o, O_p \cup \{\mathtt{const}, x_i\})$.
        
        \For{$k=1\text{ to }K $}\Comment{multiple control variable trails}
        \State draw a batch of data $T_k\gets\mathtt{GenData}(D_o)$.
        \State fitness score ${o}_k$, fitted constant values $c_k$, fitted expression $\phi_k \gets\mathtt{Optimize}(\phi, T_k)$.   
        \EndFor
        \State decide constant type for $\phi$ using $\{({o}_k, c_k)\}_{k=1}^K$. \Comment{In Section~\ref{sec:distill}}
            \State construct start symbol $S$ for next round, from $\phi$ and constant types.
        %
        
         \State fitness score ${o}_g$, fitted constant values $c_g$, fitted expression $\phi_g \gets\mathtt{Optimize}(\phi, T_g)$. 
    \State saving $\langle{o}_g, c_g, \phi_g \rangle$ into $\mathcal{Q}$.
    \EndFor
   \State \Return the equation with best fitness score in $\mathcal{Q}$.
   \Statex
   \hrulefill
   \Function{\textsc{Dpg}}{start symbol $S$, data oracle $D_o$, allowed operators and variables $O_p$}
   \State initialize $Q=$[].
   \State construct grammar rules from $O_p$ 
   \State set input and output vocabulary for RNN with the grammar rules.
   \State sets the initial input of RNN as the start symbol $S$.
     \For{$t \gets 1 \textit{ to } \mathtt{\#epochs} $}
     
   \State  sample $N$ sequences of grammar rules $\{\tau_i\}_{i=1}^N$ from RNN. \Comment{In Section~\ref{sec:rnn-train}}
   
   \State construct expressions $\{\phi_i\}_{i=1}^N$ from grammar rules  $\{\tau_i\}_{i=1}^N$.   \Comment{In Section~\ref{sec:gram-rule}}
   \For{$i=1\text{ to }N $}\Comment{optimize open constants in each expression}
        \State draw data $T_i\gets\mathtt{GenData}(D_o)$.
        \State fitness score ${o}_i$, fitted constant values $c_i$, fitted expression $\phi_i\gets\mathtt{Optimize}(\phi, T_i)$.   
        \State compute $\mathtt{reward}(\phi_i)$ using fitness score  ${o}_i$.
        \State saving $\langle{o}_i, c_i, \phi_i \rangle$ into $\mathcal{Q}$.
        \EndFor
    \State compute the estimated policy gradient $g_t\gets \frac{1}{N}\sum_{i=1}^N(\mathtt{reward}(\tau^i)-b)\nabla_{\theta}\log p(\tau^i|\theta)$.
    \State update parameters of RNN by gradient descent $\theta^t\gets \theta^{t-1}+\alpha g_t$.
    \EndFor
  \State \Return   the expression in $\mathcal{Q}$ with best fitness score.
\EndFunction
\end{algorithmic}
\end{algorithm*}

\paragraph{Objective and its Gradient} The loss function of \method is informed by the REINFORCE algorithm~\cite{DBLP:journals/ml/Williams92}, which is based on the log-derivative property: $\nabla_{\theta}p_{\theta}(x) = p_{\theta}(x) \nabla_{\theta}\log p_{\theta}(x)$, where $p_{\theta}(x)\in(0,1)$ represents a probability distribution over input $x$ with parameters $\theta$ and notation $\nabla_\theta$ is the partial derivative with respect to $\theta$.
In our formulation, let $p(\tau|\theta)$ denote the probability of sampling a sequence of grammar rules $\tau$ and $\mathtt{reward}(\tau)=1/(1+\mathtt{NMSE}(\phi))$. Here $\phi$ is the corresponding expression constructed from the rules $\tau$ following the procedure in Section~\ref{sec:gram-rule}. the probability $p(\tau|\theta)=$ is modeled by 
The learning objective is to maximize the expected reward of the sampled expressions from the RNN:
\begin{equation*}
    \arg\max_{\theta}\;\mathbb{E}_{\tau\sim p(\tau|\theta)}[\mathtt{reward}(\tau)]
\end{equation*}
Based on the REINFORCE algorithm, the gradient of the objective can be expanded as:
\begin{equation*}
\begin{aligned}
\nabla_{\theta}\mathbb E_{\tau\sim p(\tau|\theta)}[\mathtt{reward}(\tau)] & = \nabla_{\theta}\sum_{\tau\in \Sigma}{\mathtt{reward}(\tau)p(\tau|\theta) }\\
		& = \sum_{\tau\in \Sigma}{\mathtt{reward}(\tau)\nabla_{\theta}p(\tau|\theta) }\\
		& = \sum_{\tau\in \Sigma}{\mathtt{reward}(\tau)p(\tau|\theta) \nabla_{\theta}\log p(\tau|\theta)}\\
		& = \mathbb E_{\tau\sim p(\tau|\theta) }\left[\mathtt{reward}(\tau)\nabla_{\theta}\log p(\tau|\theta) \right]\\
\end{aligned}
\end{equation*}
where $\Sigma$ represents all possible sequences of grammar rules sampled from the RNN.
The above expectation can be estimated by computing the averaged over samples drawn from the distribution $p(\tau|\theta)$.
 We first sample several times from the RNN module and obtain $N$ sequences $(\tau^1,\ldots, \tau^N)$, an unbiased estimation of the gradient of the objective is computed as: $\nabla_{\theta}J(\theta)\approx \frac{1}{N}\sum_{i=1}^N\mathtt{reward}(\tau^i)\nabla_{\theta}\log p(\tau^i|\theta)$.
In practice, the above computation has a high variance. To reduce variance, it is
common to subtract a baseline function $b$ from the reward. In this study, we choose the baseline function as the average of the reward of the current sampled batch expressions. Thus we have:
 \begin{align*}
\nabla_{\theta}J(\theta)&\approx \frac{1}{N}\sum_{i=1}^N(\mathtt{reward}(\tau^i)-b)\nabla_{\theta}\log p(\tau^i|\theta), \qquad\text{ where } b=\sum_{i=1}^N \mathtt{reward}(\tau^i).
\end{align*}

Based on the description of the execution pipeline of the proposed \method, we summarize every step in Algorithm~\ref{alg:main}.
 
\paragraph{Implementation of \method}

In the experiments, we use Long short-term memory (LSTM) as the RNN layer and we configure the number of RNN layers as $3$. The dimension of the input embedding layer and the hidden vector in LSTM is configured as 512.  We use the Adam optimizer as the gradient descent algorithm with a learning rate of 0.009. The learning epoch for each round is configured 30. The maximum sequence of grammar rules is fixed to be $20$. The number of expressions sampled from the RNN is set as 1024.  When fitting the values of open constants in each expression, we sample a batch of data with batch size $1024$ from the data Oracle. The open constants in the expressions are fitted on the data using the BFGS optimizer~\footnote{\url{https://docs.scipy.org/doc/scipy/reference/optimize.minimize-bfgs.html}}. We use a multi-processor library to fit multiple expressions using 8 CPU cores in parallel. This greatly reduced the total training time.

An expression containing placeholder symbol $A$ or containing more than 20 open constants is not evaluated on the data, the fitness score of it is $-\infty$.  In terms of the reward function in the policy gradient objective, we use $\mathtt{reward}(\tau)=\frac{1}{1+\mathtt{NMSE}(\phi)}$. The normalized mean-squared error metric is further defined in Equation~\ref{eq:loss-function}.

The deep network part is implemented using the most recent TensorFlow, the expression evaluation is based on the Sympy library, and the step for fitting open constants in expression with the dataset uses the Scipy library. 
 Please find our code repository at:
\begin{mdframed}
\url{https://github.com/jiangnanhugo/VSR-DPG}
\end{mdframed}
It contains  1) the implementation of our \method method, 2) the list of datasets, and 3) the implementation of several baseline algorithms.

%% file: tex/10.expset.tex
\section{Experiment Settings}\label{apx:exp-set}
\subsection{Evaluation Metrics} \label{sec:apx-evaluation}
The goodness-of-fit indicates how well the learning algorithms perform in discovering unknown symbolic expressions.  Given a testing dataset $\mathcal{D}_{\text{test}}=\{(\mathbf{x}_{i},y_i)\}_{i=1}^n$ generated from the ground-truth expression, we measure the goodness-of-fit of a predicted expression $\phi$, by evaluating the mean-squared-error (MSE) and normalized-mean-squared-error (NMSE):
\begin{equation}\label{eq:loss-function}
\begin{aligned}
\text{MSE}&=\frac{1}{n}\sum_{i=1}^n(y_{i}-\phi(\mathbf{x}_{i}))^2, \\ 
\text{NMSE}&=\frac{\frac{1}{n}\sum_{i=1}^n(y_{i}-\phi(\mathbf{x}_{i}))^2}{\sigma_y^2}, \\
\end{aligned}
\end{equation}
The empirical variance $\sigma_y=\sqrt{\frac{1}{n}\sum_{i=1}^n \left(y_i-\frac{1}{n}\sum_{i=1}^n y_i\right)^2}$. 
We use the NMSE as the main criterion for comparison in the experiments and present the results on the remaining metrics in the case studies. 
The main reason is that the NMSE is less impacted by the output range. The output ranges of expression are dramatically different from each other, making it difficult to present results uniformly if we use other metrics.

Prior work~\cite{DBLP:conf/iclr/PetersenLMSKK21} further proposed coefficient of determination $R^2$-based Accuracy over a group of expressions in the dataset, as a statistical measure of whether the best-predicted expression is almost close to the ground-truth expression. An $R^2$ of 1 indicates that the regression predictions perfectly fit the data~\cite{nagelkerke1991note}. Given a threshold value $\mathtt{thresh}$ (we use $\mathtt{thresh}=0.9999$), for a dataset containing fitting tasks of $N$ expressions, the algorithm finds a group of best expressions $[\phi_1,\ldots, \phi_N]$ correspondingly. The $R^2$-based accuracy is computed as follows:
\begin{align*}
R^2\text{- based Accuracy}&=\frac{1}{n}\sum_{i=1}^{n}\mathbf{1}(R^2(\phi_i)\ge \mathtt{thresh}),
\end{align*}
where $R^2(\phi_i)=1-\frac{\frac{1}{n}\sum_{i=1}^n(y_{i}-{\phi}(\mathbf{x}_{i}))^2}{\sigma_y^2}$ and $\mathbf{1}(\cdot)$ is an indicator function that outputs $1$ when the $R^2(\phi_i)$ exceeds the threshold $\tau$.

\subsection{Symbolic Regression on Algebraic Equations} \label{apx:alge-eq}

\paragraph{Baselines} We consider a list of current popular baselines based on genetic programming\footnote{\url{https://github.com/jiangnanhugo/cvgp}}:
\begin{itemize}
    \item Genetic Programming (GP) maintains a population of candidate symbolic expressions, in which this population \textit{evolves} between generations. In each generation, candidate expressions undergo \textit{mutation} and \textit{crossover} with a pre-configured probability value. Then in the \textit{selection} step, expressions with the highest fitness scores (measured by the difference between the ground truth and candidate expression evaluation) are selected as the candidates for the next generation, together with a few randomly chosen expressions, to maintain diversity. After several generations, expressions with high fitness scores, \textit{i.e.}, those expressions that fit the data well survive in the pool of candidate solutions. The best expressions in all generations are recorded as {hall-of-fame} solutions.  
    \item Vertical symbolic regression with genetic programming (VSR-GP) builds on top of GP. It discovers the ground-truth expression following the vertical discovery path. In the $t$-th round, it controls variables $x_{t+1},\ldots, x_n$ as constant, and only discovers the expression involving the rest variables $x_1,\ldots, x_n$.
\end{itemize}

Eureqa~\cite{DBLP:journals/gpem/Dubcakova11} is the current best commercial software based on evolutionary search algorithms. Eureqa works by uploading the dataset $\mathcal{D}$ and the set of operators as a configuration file to its commercial server.  This algorithm is currently maintained by the DataRobot webiste\footnote{\url{https://docs.datarobot.com/en/docs/modeling/analyze-models/describe/eureqa.html}}. Computation is performed on its commercial server and only the discovered expression will be returned after several hours.  We use the provided Python API to send the training dataset to the DataRobot website and collect the predicted expression from the server-returned result. 
For the Eureqa method, the fitness measure function is negative RMSE. We generated large datasets of size $10^5$ in training each dataset.

A line of methods based on reinforcement learning\footnote{\url{https://github.com/dso-org/deep-symbolic-optimization}}:
\begin{itemize}
    \item Deep Symbolic Regression (DSR)~\cite{DBLP:conf/iclr/PetersenLMSKK21} uses a combination of recurrent neural network (RNN)  and reinforcement learning for symbolic regression. The RNN generates possible candidate expressions, and is trained with a risk-seeking policy gradient objective to generate better expressions.
    \item Priority queue training (PQT)~\cite{DBLP:journals/corr/abs-1801-03526} also uses the RNN similar to DSR for generating candidate expressions. However, the RNN is trained with a supervised learning objective over a data batch sampled from a maximum reward priority queue, focusing on optimizing the best-predicted expression.
    \item Vanilla Policy Gradient (VPG)~\cite{DBLP:journals/ml/Williams92} is similar to DSR method for the RNN part. The difference is that VPG uses the classic REINFORCE method for computing the policy gradient objective.
    
    \item Neural-Guided Genetic Programming Population Seeding (GPMeld)~\cite{DBLP:conf/nips/MundhenkLGSFP21} uses the RNN to generate candidate expressions, and these candidate expressions are improved by a genetic programming (GP) algorithm. 
\end{itemize}

\begin{table}[!t]
    \centering
    
    \begin{tabular}{r|ccc} 
    \multicolumn{4}{c}{\textbf{(a)} Genetic Programming-based methods.} \\
    \toprule
          & \cvgp & GP&Eureqa  \\\midrule
         Fitness function & NegMSE &NegMSE &NegRMSE  \\ 
         Testing set size & $256$ & $256$ & $ 50, 000$  \\
      \#CPUs for training& 1 & 1 & N/A \\\midrule
        \#genetic generations & 200 & 200  &10,000\\
      
        Mutation Probability & {0.8} & 0.8 \\
        Crossover Probability & {0.8} & 0.8   \\
        \midrule
        \multicolumn{4}{c}{\textbf{(b)} Monte Carlo Tree Search-based methods.} \\
    \toprule
         &  MCTS \\ \midrule
        Fitness function &NegMSE  \\ 
         Testing set size & $256$ \\
         \#CPUs for training & 1 \\
         \midrule 
         \multicolumn{4}{c}{\textbf{(c)} Deep reinforcement learning-based methods.} \\
         \toprule
          &  DSR & PQT & GPMeld \\\midrule
         Reward function  & \multicolumn{3}{c}{1/(1+NRMSE)} \\ 
        Training set size  & \multicolumn{3}{c}{$ 50, 000$} \\
         Testing set size& \multicolumn{3}{c}{$ 256$} \\
         Batch size & \multicolumn{3}{c}{$1024$} \\
      \#CPUs for training& \multicolumn{3}{c}{8} \\
  $\epsilon$-risk-seeking policy  & 0.02 & N/A & N/A\\ \midrule
        \#genetic generations & N/A & N/A  & 60 \\
        \#Hall of fame  & N/A& N/A& 25  \\
        Mutation Probability  & N/A &N/A  & 0.5 \\
        Crossover Probability& N/A &N/A  & 0.5   \\
        \bottomrule
    \end{tabular}
    \caption{Major hyper-parameters settings for all the algorithms considered in the experiment.}
    \label{tab:baseline-hyper-config}
\end{table}

Symbolic Physics Learner (SPL) is a heuristic search algorithm based on Monte Carlo Tree Search for finding optimal sequences of production rules using context-free grammars~\cite{DBLP:conf/icml/KamiennyLLV23,DBLP:conf/iclr/Sun0W023}\footnote{\url{https://github.com/isds-neu/SymbolicPhysicsLearner}}. 
It employs Monte Carlo simulations to explore the search space of all the production rules and determine the value of each node in the search tree. SPL consists of four steps in each iteration:   1) Selection. Starting at a root node, recursively select the optimal child (\textit{i.e.}, one of the production rules) until reaching an expandable node or a leaf node. 2) Expansion. If the expandable node is not the terminal, create one or more of its child nodes to expand the search tree. 3) Simulation. Run a simulation from the new node until achieving the result. 4) Backpropagation. Update the node sequence from the new node to the root node with the simulated result. To balance the selection of optimal child node(s) by exploiting known rewards (exploitation) or expanding a new node to explore potential rewards exploration, the upper confidence bound (UCB) is often used.

End to End Transformer for symbolic regression (E2ETransformer)~\cite{DBLP:conf/nips/KamiennydLC22}\footnote{\url{https://github.com/facebookresearch/symbolicregression}}. 
They propose to use a deep transformer to pre-train on a large set of randomly generated expressions. We load the shared pre-trained model. We provide the given dataset and the E2ETransformer infers 10 best expressions. We choose to report the expression with the best NMSE scores.

We list the major hyper-parameter settings for all the algorithms in Table~\ref{tab:baseline-hyper-config}. Note that if we use the default parameter settings, the GPMeld algorithm takes more than 1 day to train on one dataset. 
Because of such slow performance, we cut the number of genetic programming generations in GPMeld by half to ensure fair comparisons with other approaches.

\paragraph{Dataset for Algebraic Equations} The dataset is available at the code repository with the folder name:\\
$\small{\mathtt{data/algebraic\_equations/equations\_trigonometric}}$.\\
The expressions used for comparison have the same mathematical operators $O_p=\{+,-,\times, \sin,\cos\}$. One configuration $(2,1,1)$ is shown in Table~\ref{tab:sincos211-data}.

\begin{table}[!t]
    \centering\small
    \begin{tabular}{c|l}
    \hline
        Eq. ID & Exact Expression  \\ \hline
        prog-0 &  $-0.167\sin(x_0) \cos(x_1) + 0.4467 \cos(x_0) - 0.2736$ \\
        prog-1 & $0.6738 x_0 - 0.5057\sin(x_0)\sin(x_1) + 0.8987$ \\
        prog-2 & $-0.5784 x_0 x_1 + 0.556 \cos(x_1) + 0.8266$ \\
        prog-3 & $0.0882x_0 - 0.7944\sin(x_0)\sin(x_1) + 0.4847$ \\
        prog-4 & $-0.7262\sin(x_1)\cos(x_0) - 0.006\cos(x_1) - 0.9218$\\
        prog-5 & $0.189x_0x_1 - 0.7125\cos(x_1) - 0.4207$ \\
        prog-6 &  $0.2589x_0\sin(x_1) + 0.1977x_1 - 0.7504$\\
        prog-7 & $-0.2729x_0\sin(x_1) - 0.7014x_1 + 0.3248$\\
        prog-8 & $-0.2582x_0 - 0.8355x_1\cos(x_0) - 0.5898$\\
        prog-9 & $0.1052x_0x_1 + 0.0321x_0 - 0.9554$\\
    \hline
    \end{tabular}
    \caption{10 randomly drawn expressions with 2 variables, 1 single term, and 1 cross term with operators $\{\sin,\cos,+,-,\times\}$.}
    \label{tab:sincos211-data}
\end{table}

For the extended analysis, where we consider many more input variables, they are available in the folder with the name:
\begin{center}
$\small{\mathtt{data/algebraic\_equations/large\_scale\_}n}$
\end{center}
where the value of $n$ is the number of total variables in, which can be $10, 20,30,40,50$.

The original expression is: $-0.4156x_0x_1 - 0.1399x_2\cos(x_1) + 0.0438x_2 + 0.9508x_3\sin(x_1) + 0.2319x_3 - 0.6808x_4\cos(x_3) - 0.4468x_4 + 0.0585\sin(x_0) + 0.6224\cos(x_1) - 0.8638\cos(x_2)\cos(x_3) + 0.959$.
We extend this expression by choosing 5 variables from the total $n=10$ variables and mapping the selected variables to the variables $x_0,\ldots,x_4$. Here are 10 randomly generated expressions:
\begin{align*}
\phi_1=&- 0.4156x_3x_9 - 0.1399x_1\cos(x_3) + 0.0438x_1+ 0.9508x_0\sin(x_3)  + 0.2319x_0   - 0.6808x_4\cos(x_0)  \\
&- 0.4468x_4+ 0.0585\sin(x_9)  + 0.6224\cos(x_3) - 0.8638\cos(x_0)\cos(x_1)+ 0.959 \\
\phi_2=&-0.4156x_0x_5 - 0.1399x_3\cos(x_0) + 0.0438x_3+ 0.9508x_1\sin(x_0) + 0.2319x_1  - 0.6808x_7\cos(x_1) \\
&- 0.4468x_7 + 0.0585\sin(x_5) + 0.6224\cos(x_0) - 0.8638\cos(x_1)\cos(x_3) + 0.959\\
\phi_3=& - 0.4156x_5x_8- 0.1399x_1\cos(x_5) + 0.0438x_1 + 0.9508x_4\sin(x_5) + 0.2319x_4 -0.6808x_0\cos(x_4) \\
& - 0.4468x_0+ 0.0585\sin(x_8)  + 0.6224\cos(x_5) - 0.8638\cos(x_1)\cos(x_4)+ 0.959\\
\phi_3=&-0.4156x_2x_6 - 0.1399x_3\cos(x_2) + 0.0438x_3 + 0.9508x_7\sin(x_2) + 0.2319x_7 - 0.6808x_9\cos(x_7) \\
&- 0.4468x_9 + 0.0585\sin(x_6) + 0.6224\cos(x_2) - 0.8638\cos(x_3)\cos(x_7) + 0.959\\
\phi_4=&- 0.4156x_3x_7  - 0.1399x_8\cos(x_3)  + 0.0438x_8  +0.9508x_2\sin(x_3) + 0.2319x_2 - 0.6808x_9\cos(x_2) \\
&- 0.4468x_9 + 0.0585\sin(x_7)  + 0.6224\cos(x_3) - 0.8638\cos(x_2)\cos(x_8)+ 0.959 \\
\phi_5=&- 0.4156x_1x_3 - 0.1399x_6\cos(x_3) + 0.0438x_6+ 0.9508x_2\sin(x_3) + 0.2319x_2 -0.6808x_0\cos(x_2) \\
& - 0.4468x_0 + 0.0585\sin(x_1)  + 0.6224\cos(x_3) - 0.8638\cos(x_2)\cos(x_6) + 0.959\\
\phi_6=&-0.4156x_4x_5 - 0.1399x_7\cos(x_5)+ 0.0438x_7 + 0.9508x_6\sin(x_5) + 0.2319x_6  - 0.6808x_8\cos(x_6) \\
&- 0.4468x_8 + 0.0585\sin(x_4) + 0.6224\cos(x_5) - 0.8638\cos(x_6)\cos(x_7) + 0.959\\
\phi_7=&  - 0.4156x_3x_8- 0.1399x_5\cos(x_3) + 0.0438x_5 0.9508x_0\sin(x_3)+ 0.2319x_0 - 0.6808x_7\cos(x_0) \\
&- 0.4468x_7 + 0.0585\sin(x_8)  + 0.6224\cos(x_3) - 0.8638\cos(x_0)\cos(x_5) + 0.959\\
\phi_8=&-0.4156x_0x_3 - 0.1399x_2\cos(x_0) + 0.0438x_2 + 0.9508x_5\sin(x_0) + 0.2319x_5 - 0.6808x_6\cos(x_5) \\
&- 0.4468x_6 + 0.0585\sin(x_3) + 0.6224\cos(x_0) - 0.8638\cos(x_2)\cos(x_5) + 0.959 \\
\phi_9=&-0.4156x_0x_5 - 0.1399x_8\cos(x_5)+ 0.0438x_8- 0.6808x_2\cos(x_7)  + 0.9508x_7\sin(x_5) + 0.2319x_7  \\
& - 0.4468x_2+ 0.0585\sin(x_0) + 0.6224\cos(x_5) - 0.8638\cos(x_7)\cos(x_8) + 0.959
\end{align*}
The rest expressions are available in the folder.

\subsection{Symbolic Regression on Ordinary Differential Equations} \label{apx:diff-eq}
 The temporal evolution of the system is modeled by the time derivatives of the state variables. Let $\mathbf{x}$ be the $n$-dimensional vector of state variables, and $\diff \mathbf{x}/ \diff t$ is the vector of their time derivatives, which is noted as $\dot{\mathbf{x}}$ for abbreviation. The ordinary differential equation (ODEs) is of the form $\dot{\mathbf{x}}=\phi(\mathbf{x},\mathbf{c})$, where constant vector $\mathbf{c}\in\mathbb{R}^m$ are parameters of the ODE model. Given the initial state $\mathbf{x}(t_0)$,  the finite time difference $\Delta t$ and the expression $\phi(\mathbf{x},\mathbf{c})$, the ODEs are numerically simulated to obtain the state trajectory $\mathbf{x}(t_1), \ldots, \mathbf{x}(t_N)$, where $\mathbf{x}(t_i)=\mathbf{x}(t_{i-1})+\phi(\mathbf{x},\mathbf{c})\Delta t$ and $t_i=t_{i-1}+\Delta t$. 

\paragraph{Task Definition}
Following the definition of symbolic regression on differential equation in~\cite{DBLP:conf/dis/GecOBDT22,DBLP:conf/iclr/Sun0W023},
given a trajectory dataset of state variable and its time derivatives $\{(\mathbf{x}(t_i),\dot{\mathbf{x}}(t_i))\}_{i=1}^N$, $\dot{\mathbf{x}}(t_i)$ represents the value
of the derivative of variable $\mathbf{x}$ at time $t_i$,  the symbolic regression task is to predict the best expression $\phi(\mathbf{x},\mathbf{c})$ that minimizes the average loss on trajectory data:
\begin{equation*}
\arg\min_{\phi}\frac{1}{N}\sum_{i=1}^N\ell(\dot{\mathbf{x}}(t_i),\phi(\mathbf{x}(t_i),\mathbf{c}))
\end{equation*}
Other formulations of this problem assume we have no access to its time derivatives, that is $\{(t_i, \mathbf{x}(t_i))\}_{i=1}^N$~\cite{2023odeformer}. This formulation is tightly connected to our setting and relatively more challenging. We can still estimate the finite difference between the current and next state variables as its approximated time derivative: $\dot{\mathbf{x}}(t_i)=\frac{\mathbf{x}(t_i)-\mathbf{x}(t_{i-1})}{t_i-t_{i-1}}$.

\paragraph{Baselines} For the baselines on the differentiable equations, we consider
\begin{itemize}
\item SINDy~\cite{brunton2016sparse}\footnote{\url{https://github.com/dynamicslab/pysindy}} is a popular method using a sparse regression algorithm to find the differential equations.
\item ODEFormer~\cite{2023odeformer}\footnote{\url{https://github.com/sdascoli/odeformer}} is the most recent framework that uses the transformer for the discovery of ordinary differential equations. We use the provided pre-trained model to predict the governing expression with the dataset. We execute the model 10 times and pick the expression with the smallest NMSE error. The dataset size is $500$, which is the largest dataset configuration for the ODEFormer.
    \item ProGED~\cite{DBLP:journals/kbs/BrenceTD21}\footnote{\url{https://github.com/brencej/ProGED}} uses probabilistic context-free grammar to search for differential equations. ProGED first samples a list of candidate expressions from the defined probabilistic context-free grammar for symbolic expressions. Then ProGED fits the open constants in each expression using the given training dataset. The equation with the best fitness scores is returned.
\end{itemize}

\paragraph{Dataset for Differential Equations.}
We collect a set of real-world ordinary differential equations of multiple input variables from the SINDy codebase\footnote{\url{https://github.com/dynamicslab/pysindy/blob/master/pysindy/utils/odes.py}}. 

\begin{itemize}
    \item {Lorenz Attractor.} Let $x_0,x_1,x_2$ be functions of time $x_0(t), x_1(t), x_2(t)$ and stands for the position in the $(x, y, z)$ coordinates. 
Here we consider 3-dimensional Lorenz system whose dynamical behavior $(x_0, x_1, x_2)$ is governed by
\begin{equation*}
\begin{aligned}
\dot x_0  &= \sigma (x_1 - x_0), \\
\dot x_1  &= x_0(\rho - x_2) - x_1,\\ 
\dot x_2 &= x_0x_1 - \beta x_2,
\end{aligned}
\end{equation*}
 with parameters $\sigma = 10, \beta = 8/3, \rho = 28$. 

\item {Glycolysis Oscillations}.  The dynamic behavior of yeast glycolysis can be described as a set of 7 variables $x_0,\ldots,x_6$. The biological definition of each variable from ~\citeauthor{brechmann2021unbounded} is provided in Table~\ref{tab:Gly-var-def}. The governing equations are:
\begin{equation*}
\begin{aligned}
    \dot x_0 &= J_0 - \frac{(k_1x_0x_5)}{(1 + (x_5 / K_1) ^ q)},\\
    \dot x_1 &=\frac{2(k_1x_0x_5)}{1 + (x_5 / K_1) ^ q} - k_2x_1(N - x_4) - k_6x_1x_4, \\
    \dot x_2 &=k_2x_1(N - x_4) - k_3x_2(A - x_5), \\
    \dot x_3 &=k_3x_2(A - x_5) - k_4x_3x_4 - \kappa(x_3 - x_6), \\
    \dot x_4 &=k_2x_1(N - x_4) - k_4x_3x_4 - k_6x_1x_4, \\
    \dot x_5 &=\frac{-2k_1x_0x_5}{1 + (x_5 / K_1) ^ q} + 2k_3 x_2(A - x_5) - k_5x_5, \\
    \dot x_6 &=\phi\kappa(x_3 - x_6) - K x_6
\end{aligned}
\end{equation*}
where the parameters $J_0 = 2.5, k_1 = 100, k_2 = 6, k_3 = 16, k_4 = 100, k_5 = 1.28, k_6 = 12, K = 1.8, \kappa = 13, q = 4, K_1 = 0.52, \phi = 0.1, N = 1, A = 4$.  The rest of the differential equations from this Glycolysis family can be found at~\cite{brechmann2021unbounded}.
\begin{table*}[!t]
    \centering
    \begin{tabular}{c|l|c|c}
\hline
     Variable & Biological Definition & Range& Standard deviation \\ \hline
$x_0$ & Glucose  & $[0.15, 1.60]$ & $0.4872$ \\
$x_1$ & Glyceraldehydes-3-phosphate  & \multirow{2}{*}{$[0.19, 2.16]$} & \multirow{2}{*}{$0.6263$} \\
& and dihydroxyacetone phosphate pool & &\\
$x_2$ & 1,3-bisphosphoglycerate & $[0.04, 0.20]$ & $0.0503$ \\
$x_3$ & Cytosolic pyruvate and acetaldehyde pool & $[0.10, 0.35]$ & $0.0814$ \\
$x_4$ & NADH & $[0.08, 0.30]$ & $0.0379$ \\
$x_5$ & ATP  & $[0.14, 2.67]$ & $0.7478$ \\
$x_6$  & Extracellular pyruvate and acetaldehyde pool & $[0.05, 0.10]$ & $0.0159$ \\
\hline 
    \end{tabular}
    \caption{Biological definition of variables in Glycolysis Oscillations. The allowed range of initial states for the training data set and the standard deviation of the limit cycle are also included.}
    \label{tab:Gly-var-def}
\end{table*}
\item MHD turbulence. The following equations describe the dynamic behavior of the  Carbone and Veltri triadic MHD model:
\begin{equation*}
\begin{aligned}
\dot x_0 &=-2 \nu x_0 + 4 (x_1 x_2 - x_4 x_5), \\
\dot x_1 &= -5 \nu x_1 - 7 (x_0 x_2 - x_3 x_5),\\ 
\dot x_2 &= -9 \nu x_2 + 3 (x_0 x_1 - x_3 x_4), \\
\dot x_3 &=  -2 \mu x_4 + 2 (x_5 x_1 - x_2 x_4),\\
\dot x_4 &=  -5 \mu x_4 + \sigma x_5 + 5 (x_2 x_3 - x_0 x_5),\\
\dot x_5 &=   -9 \mu x_5 + \sigma x_4 + 9 (x_4 x_0 - x_1 x_3),
\end{aligned}
\end{equation*}
where the parameters $\nu=0, \mu=0,\sigma=0$. \cite{de2006role} define  $x_0,x_1,x_2$ as the velocity and $x_3,x_4,x_5$ as to the magnetic field. $\nu,\mu$ represents, respectively, the kinematic viscosity and the resistivity.
\end{itemize}
We notice there is a recently proposed dataset ODEBench~\cite{2023odeformer}. It is not selected for study, since it mainly contains differential equations up to two variables.

\paragraph{Evaluation Metrics.} We use the $R^2$-based Accuracy metric to evaluate if the whole set of predicted expressions has a $R^2$ score higher than $0.9999$.

%% file: tex/10.expextra.tex
\section{Extra Experiments}
\subsection{Discovered Algebraic Equations by each learning algorithm.} \label{apx:exact-result}

The predicted expression by \method (ours) for configuration $(4,4, 6)$. 60\% of the predicted expression has a $\leq10^{-6}$ NMSE score. 

The predicted result for prog-0:
\begin{align*}
 & -0.3012000175544417x_0x_3 - 0.23479995033497178x_0 +\\
 &0.045433905730119135x_1 + 0.10966141816565093x_2 + \\
 &0.22430013864298073x_3 + 0.9739999857983681\sin(x_2) +\\
 &0.3581998363171518\cos(x_2)\cos(x_3) + \\
 &0.2862218136669438\cos(x_3) + 3.126115887545009 \\
\end{align*}
The predicted result for prog-1:
\begin{align*}
  & -0.5807073848480102x_0 - 0.09660000567273663x_1 -\\
  & 0.9748000148040502x_2x_3 - 0.4638000163793846x_3\cos(x_0) \\
  &- 0.4221638801953578x_3 - 0.012754904995223835\sin(x_2) \\
  &+ 0.15999997730356633\cos(x_2) + 0.2524999760074076\cos(x_3)\\
  &+ 0.3830840657508305 \\
\end{align*}
The predicted result for prog-2:
\begin{align*}
  &0.5974706919691478x_0 + 0.8783029159486363x_1 +\\
  &0.584599994337829x_2\cos(x_1) - 0.8430097368938334x_3 - \\
  &0.4739999968689642\sin(x_2) - 3.3558075600032683e-8\sin(x_3)\\
  &- 0.3244634752208093\cos(x_2) + 0.5068000094901586\cos(x_3) \\
  &- 0.787302540612925 \\
\end{align*}
The predicted result for prog-3:
\begin{align*}
   & -0.89730000849859939x_0 - 7.242399512792391x_1 - \\
   &1.2513833693643626\cos(x_0) - 1.5175517989615754 \\
  &0.032734568821399544x_0 + 0.928299994219054x_1sin(x_0)\\
  &+ 0.11740000072851x_1 - 1.6081211674938465x_2 +\\
  &0.5674704740296996x_3 + 0.1769999997564291\sin(x_2)\cos(x_3)\\
  &- 0.4200518345694358 \\
\end{align*}
The predicted result for prog-4:
\begin{align*}
  & 0.10499999077952751x_0\sin(x_2) + 0.8918999999268585x_3\sin(x_1)\\
  &+ 0.11399999910836027x_3\cos(x_2) - 0.38250000586131516x_3 \\
  &- 0.14609999633658752x_4\sin(x_0) + 0.9090999941858626x_4\sin(x_1) \\
  &- 0.6846999999068688\sin(x_0) + 0.9993000283241971\sin(x_2) \\
  &- 0.19519999212829273\cos(x_1) + 0.6172999945789425\cos(x_4) \\
  &- 0.4587999974860775  \\
\end{align*}
The predicted result for prog-5:
\begin{align*}
  &0.3900029487047949x_0 + 0.15013453625258577x_2 + 0.7973748097464934sin(x_2) \\
  &+ 0.6004443541983869cos(x_1) + 1.4041023040405819 \\
\end{align*}

\subsection{Discovered differential equations by each Learning Algorithm.} \label{apx:tab:diff-eq-exact}

\paragraph{MHD turbulence} We collect the best-predicted expression by each algorithm.

SINDy.
\begin{align*}
\dot x_0 &= 0.195+ 0.009x_0 + 0.025 x_1 + 0.045 x_2 + 0.001 x_4 -0.012 x_5 -3.772x_0^2 -0.002x_0 x_2 \\
&+ 1.157x_0 x_3 + 0.002x_0 x_4 -0.011x_0 x_5 -2.016 x_1^2 + 3.976 x_1 x_2 -0.001 x_1 x_3 + 1.158 x_1 x_4\\
& + 0.003 x_1 x_5 + 0.306 x_2^2 -0.005 x_2 x_3 -0.007 x_2 x_4 + 1.164 x_2 x_5 + 0.602 x_3^2 \\
&+ 0.011 x_3 x_5 + 0.437 x_4^2 -3.996 x_4 x_5 -1.426 x_5^2 \\
\dot x_1 &= -1.046+ 0.011x_0 -0.008 x_1 + 0.01 x_2 -0.005 x_3 -0.003 x_4 -0.012 x_5 -0.686x_0^2 \\
&+ 0.007x_0 x_1 -7.015x_0 x_2 + 0.030x_0 x_3 -0.004x_0 x_4 + 0.011x_0 x_5 + 0.075 x_1^2 + 0.013 x_1 x_2 \\
&-0.003 x_1 x_3 + 0.035 x_1 x_4+ 0.001 x_1 x_5 + 1.108 x_2^2 + 0.010 x_2 x_3 + 0.003 x_2 x_4+ 0.043 x_2 x_5\\
& -0.370 x_3^2 + 0.001 x_3 x_4 + 6.997 x_3 x_5+ 0.518 x_4^2 + 0.005 x_4 x_5 -0.015 x_5^2\\
\end{align*}
\begin{align*}
\dot x_2 &= 0.098+ 0.003x_0 -0.007 x_1 -0.007 x_2 -0.007 x_3 + 0.002 x_5 -0.965x_0^2 + 2.993x_0 x_1 \\
&-0.004x_0 x_2 + 0.248x_0 x_3 + 0.002x_0 x_5 -0.582 x_1^2 + 0.007 x_1 x_2+ 0.255 x_1 x_4 + 0.001 x_1 x_5\\
& -0.050 x_2^2 + 0.008 x_2 x_3 + 0.001 x_2 x_4 + 0.248 x_2 x_5 + 0.486 x_3^2 -2.997 x_3 x_4-0.001 x_3 x_5 \\
&+ 0.161 x_4^2 -0.002 x_4 x_5 -0.340 x_5^2\\
\dot x_3 &= -0.027+ 0.004x_0 -0.003 x_1 -0.012 x_2 + 0.001 x_3+ 0.001 x_4 + 0.002 x_5 -2.958x_0^2 \\
&-0.013x_0 x_2 + 0.750x_0 x_3 + 0.019x_0 x_5 -1.610 x_1^2 + 0.009 x_1 x_2 -0.003 x_1 x_3 + 0.751 x_1 x_4 \\
&+ 1.986 x_1 x_5 + 0.198 x_2^2 + 0.007 x_2 x_3 -2.004 x_2 x_4 + 0.749 x_2 x_5 + 1.185 x_3^2-0.013 x_3 x_5 \\
&+ 0.589 x_4^2 + 0.005 x_4 x_5 -0.989 x_5^2\\
\dot x_4 &= -0.434+ 0.024x_0 + 0.008 x_1 -0.001 x_2 -0.002 x_3-0.006 x_4 -0.015 x_5 + 3.462x_0^2 \\
&+ 0.002x_0 x_1 -0.039x_0 x_2 -1.182x_0 x_3 -0.005x_0 x_4 -4.975x_0 x_5 + 2.168 x_1^2 -0.019 x_1 x_2 \\
&-1.179 x_1 x_4 + 0.033 x_1 x_5 + 0.455 x_2^2 + 5.005 x_2 x_3 + 0.018 x_2 x_4 -1.162 x_2 x_5 -1.890 x_3^2 \\
&-0.012 x_3 x_5 -0.482 x_4^2 -0.009 x_4 x_5 + 1.269 x_5^2\\
\dot x_5 &= -1.775-0.015x_0 -0.022 x_1 + 0.121 x_2 -0.032 x_3 + 0.009 x_4 -0.035 x_5 + 21.145x_0^2 \\
&+ 0.013x_0 x_1 + 0.016x_0 x_2 -5.838x_0 x_3 + 8.978x_0 x_4 + 0.010x_0 x_5 + 11.874 x_1^2 + 0.023 x_1 x_2 \\
& -8.993 x_1 x_3 -5.863 x_1 x_4 -0.580 x_2^2 + 0.028 x_2 x_3 -0.008 x_2 x_4 -5.810 x_2 x_5  -6.541 x_3^2 \\
&+ 0.003 x_3 x_4 + 0.004 x_3 x_5 -2.911 x_4^2 -0.003 x_4 x_5 + 7.768 x_5^2\\
\end{align*}
 ODEFormer 
        \begin{equation*}
        \begin{aligned}
            \dot x_0 &= 0.0093  x_0  (-0.1332 - x_2)^2 \\
\dot x_1 &= -4.8118  x_2 \\
\dot x_2 &= 2.4147  x_1 -1.3145  \sin(-0.1171 + 15.0423  x_1) \\
\dot x_3 &= -3.6859  x_1  x_2 \\
\dot x_4 &= 0.9808  x_2 -3.7675  x_5 \\
\dot x_5 &= \frac{0.0105}{ -11.23 + 7.7065  x_2} + 8.2969  x_4-2.2755  x_1 \\
        \end{aligned}
        \end{equation*}

SPL
\begin{align*}
\dot x_0 &=-0.2x_0 + 4x_1x_2 - 4x_4x_5\\
\dot x_1 &=-7x_0x_2 - 0.5x_1 + 6.99x_3x_5\\
\dot x_2 &=2.95x_0x_1 - 3.02x_3x_4 \\
\dot x_3 &=-2.07x_2x_4 + 0.435 \\
\dot x_4 &=-4.97x_0x_5 + 5.0x_2x_3 + 0.045x_2x_5 + 0.025x_3 + 0.032x_4x_5 - 0.993x_4 \\
\dot x_5 &=9.076x_0x_4 - 0.0116x_0 - 8.996x_1x_3 - 1.758x_5
\end{align*}
\method (ours) 
\begin{align*}
\dot x_0 &=-0.2x_0 + 4.0x_1x_2 - 4.0x_4x_5 \\
\dot x_1 &=-7.0x_0x_2 - 0.5x_1 + 7.0x_3x_5 \\
\dot x_2 &=3.0x_0x_1 - 0.9x_2 - 3.0x_3x_4 \\
\dot x_3 &=2.x_1x_5 - 2.x_2x_4 - 0.40x_4 \\
\dot x_4 &=-5.0x_0x_5 + 5.0x_2x_3 - 1.0x_4 + 0.3x_5 \\
\dot x_5 &=9.0x_0x_4 - 9.0x_1x_3 + 0.3x_4 - 1.8x_5
\end{align*}





%% file: arxiv.bbl
\begin{thebibliography}{56}
\providecommand{\natexlab}[1]{#1}
\providecommand{\url}[1]{\texttt{#1}}
\expandafter\ifx\csname urlstyle\endcsname\relax
  \providecommand{\doi}[1]{doi: #1}\else
  \providecommand{\doi}{doi: \begingroup \urlstyle{rm}\Url}\fi

\bibitem[Langley et~al.(1987)Langley, Simon, Bradshaw, and
  Zytkow]{langey1988scientificdiscovery}
Patrick~W. Langley, Herbert~A. Simon, Gary Bradshaw, and Jan~M. Zytkow.
\newblock \emph{Scientific Discovery: Computational Explorations of the
  Creative Process}.
\newblock The MIT Press, 02 1987.

\bibitem[Kulkarni and Simon(1988)]{kulkarni1988processes}
Deepak Kulkarni and Herbert~A Simon.
\newblock The processes of scientific discovery: The strategy of
  experimentation.
\newblock \emph{Cogn. Sci.}, 12\penalty0 (2):\penalty0 139--175, 1988.

\bibitem[Wang et~al.(2023)Wang, Fu, Du, Gao, Huang, Liu, Chandak, Liu,
  Van~Katwyk, Deac, et~al.]{wang2023scientific}
Hanchen Wang, Tianfan Fu, Yuanqi Du, Wenhao Gao, Kexin Huang, Ziming Liu, Payal
  Chandak, Shengchao Liu, Peter Van~Katwyk, Andreea Deac, et~al.
\newblock Scientific discovery in the age of artificial intelligence.
\newblock \emph{Nature}, 620\penalty0 (7972):\penalty0 47--60, 2023.

\bibitem[Schmidt and Lipson(2009)]{doi:10.1126/science.1165893}
Michael Schmidt and Hod Lipson.
\newblock Distilling free-form natural laws from experimental data.
\newblock \emph{Science}, 324\penalty0 (5923):\penalty0 81--85, 2009.

\bibitem[Lenat(1977)]{LENAT1977ubiquity}
Douglas~B. Lenat.
\newblock The ubiquity of discovery.
\newblock \emph{Artif. Intell.}, 9\penalty0 (3):\penalty0 257--285, 1977.

\bibitem[Virgolin et~al.(2019)Virgolin, Alderliesten, and
  Bosman]{DBLP:conf/gecco/VirgolinAB19}
Marco Virgolin, Tanja Alderliesten, and Peter A.~N. Bosman.
\newblock Linear scaling with and within semantic backpropagation-based genetic
  programming for symbolic regression.
\newblock In \emph{{GECCO}}, pages 1084--1092, 2019.

\bibitem[Todorovski and Dzeroski(1997)]{DBLP:conf/icml/TodorovskiD97}
Ljupco Todorovski and Saso Dzeroski.
\newblock Declarative bias in equation discovery.
\newblock In \emph{{ICML}}, pages 376--384. Morgan Kaufmann, 1997.

\bibitem[Sun et~al.(2023)Sun, Liu, Wang, and Sun]{DBLP:conf/iclr/Sun0W023}
Fangzheng Sun, Yang Liu, Jian{-}Xun Wang, and Hao Sun.
\newblock Symbolic physics learner: Discovering governing equations via monte
  carlo tree search.
\newblock In \emph{{ICLR}}, 2023.

\bibitem[Kamienny et~al.(2023)Kamienny, Lample, Lamprier, and
  Virgolin]{DBLP:conf/icml/KamiennyLLV23}
Pierre{-}Alexandre Kamienny, Guillaume Lample, Sylvain Lamprier, and Marco
  Virgolin.
\newblock Deep generative symbolic regression with monte-carlo-tree-search.
\newblock In \emph{{ICML}}, volume 202. {PMLR}, 2023.

\bibitem[Petersen et~al.(2021)Petersen, Landajuela, Mundhenk, Santiago, Kim,
  and Kim]{DBLP:conf/iclr/PetersenLMSKK21}
Brenden~K. Petersen, Mikel Landajuela, T.~Nathan Mundhenk, Cl{\'{a}}udio~Prata
  Santiago, Sookyung Kim, and Joanne~Taery Kim.
\newblock Deep symbolic regression: Recovering mathematical expressions from
  data via risk-seeking policy gradients.
\newblock In \emph{{ICLR}}, 2021.

\bibitem[Mundhenk et~al.(2021)Mundhenk, Landajuela, Glatt, Santiago, Faissol,
  and Petersen]{DBLP:conf/nips/MundhenkLGSFP21}
T.~Nathan Mundhenk, Mikel Landajuela, Ruben Glatt, Cl{\'{a}}udio~P. Santiago,
  Daniel~M. Faissol, and Brenden~K. Petersen.
\newblock Symbolic regression via deep reinforcement learning enhanced genetic
  programming seeding.
\newblock In \emph{NeurIPS}, pages 24912--24923, 2021.

\bibitem[Jiang and Xue(2023)]{DBLP:conf/pkdd/JiangX23}
Nan Jiang and Yexiang Xue.
\newblock Symbolic regression via control variable genetic programming.
\newblock In \emph{{ECML/PKDD}}, volume 14172 of \emph{Lecture Notes in
  Computer Science}, pages 178--195. Springer, 2023.

\bibitem[Jiang et~al.(2023)Jiang, Nasim, and Xue]{jiang2023vertical}
Nan Jiang, Md~Nasim, and Yexiang Xue.
\newblock Vertical symbolic regression.
\newblock \emph{arXiv preprint arXiv:2312.11955}, 2023.

\bibitem[Joule(1843)]{joule1843production}
James~Prescott Joule.
\newblock On the production of heat by voltaic electricity.
\newblock In \emph{Abstracts of the Papers Printed in the Philosophical
  Transactions of the Royal Society of London}, pages 280--282, 1843.

\bibitem[Virgolin and Pissis(2022)]{journal/tmlr/virgolin2022}
Marco Virgolin and Solon~P Pissis.
\newblock Symbolic regression is {NP}-hard.
\newblock \emph{TMLR}, 2022.

\bibitem[Abolafia et~al.(2018)Abolafia, Norouzi, and
  Le]{DBLP:journals/corr/abs-1801-03526}
Daniel~A. Abolafia, Mohammad Norouzi, and Quoc~V. Le.
\newblock Neural program synthesis with priority queue training.
\newblock \emph{CoRR}, abs/1801.03526, 2018.

\bibitem[Lehman et~al.(2004)Lehman, Santner, and Notz]{lehman2004designing}
Jeffrey~S Lehman, Thomas~J Santner, and William~I Notz.
\newblock Designing computer experiments to determine robust control variables.
\newblock \emph{Statistica Sinica}, pages 571--590, 2004.

\bibitem[Wierstra et~al.(2010)Wierstra, F{\"{o}}rster, Peters, and
  Schmidhuber]{DBLP:journals/igpl/WierstraFPS10}
Daan Wierstra, Alexander F{\"{o}}rster, Jan Peters, and J{\"{u}}rgen
  Schmidhuber.
\newblock Recurrent policy gradients.
\newblock \emph{Log. J. {IGPL}}, 18\penalty0 (5):\penalty0 620--634, 2010.

\bibitem[Williams(1992)]{DBLP:journals/ml/Williams92}
Ronald~J. Williams.
\newblock Simple statistical gradient-following algorithms for connectionist
  reinforcement learning.
\newblock \emph{Mach. Learn.}, 8:\penalty0 229--256, 1992.

\bibitem[Fletcher(2000)]{fletcher2000practical}
Roger Fletcher.
\newblock \emph{Practical methods of optimization}.
\newblock John Wiley \& Sons, 2000.

\bibitem[Kirkpatrick et~al.(2021)Kirkpatrick, McMorrow, Turban, Gaunt, Spencer,
  Matthews, Obika, Thiry, Fortunato, Pfau, Castellanos, Petersen, Nelson,
  Kohli, Mori-Sánchez, Hassabis, and Cohen]{doi:10.1126/science.abj6511}
James Kirkpatrick, Brendan McMorrow, David H.~P. Turban, Alexander~L. Gaunt,
  James~S. Spencer, Alexander G. D.~G. Matthews, Annette Obika, Louis Thiry,
  Meire Fortunato, David Pfau, Lara~Román Castellanos, Stig Petersen,
  Alexander W.~R. Nelson, Pushmeet Kohli, Paula Mori-Sánchez, Demis Hassabis,
  and Aron~J. Cohen.
\newblock Pushing the frontiers of density functionals by solving the
  fractional electron problem.
\newblock \emph{Science}, 374\penalty0 (6573):\penalty0 1385--1389, 2021.

\bibitem[Jumper et~al.(2021)Jumper, Evans, Pritzel, Green, Figurnov,
  Ronneberger, Tunyasuvunakool, Bates, {\v{Z}}{\'\i}dek, Potapenko,
  et~al.]{jumper2021highly}
John Jumper, Richard Evans, Alexander Pritzel, Tim Green, Michael Figurnov,
  Olaf Ronneberger, Kathryn Tunyasuvunakool, Russ Bates, Augustin
  {\v{Z}}{\'\i}dek, Anna Potapenko, et~al.
\newblock Highly accurate protein structure prediction with alphafold.
\newblock \emph{Nature}, 596\penalty0 (7873):\penalty0 583--589, 2021.

\bibitem[Bradley et~al.(2001)Bradley, Easley, and Stolle]{BRADLEY2001reasoning}
Elizabeth Bradley, Matthew Easley, and Reinhard Stolle.
\newblock Reasoning about nonlinear system identification.
\newblock \emph{Artif. Intell.}, 133\penalty0 (1):\penalty0 139--188, 2001.

\bibitem[Dzeroski and Todorovski(1995)]{Dzeroski1995lagrange}
Saso Dzeroski and Ljupco Todorovski.
\newblock Discovering dynamics: From inductive logic programming to machine
  discovery.
\newblock \emph{J. Intell. Inf. Syst.}, 4\penalty0 (1):\penalty0 89--108, 1995.

\bibitem[Brunton et~al.(2016)Brunton, Proctor, and Kutz]{brunton2016sparse}
Steven~L. Brunton, Joshua~L. Proctor, and J.~Nathan Kutz.
\newblock Discovering governing equations from data by sparse identification of
  nonlinear dynamical systems.
\newblock \emph{{PNAS}}, 113\penalty0 (15):\penalty0 3932--3937, 2016.

\bibitem[Wu and Tegmark(2019)]{PhysRevE.100.033311}
Tailin Wu and Max Tegmark.
\newblock Toward an artificial intelligence physicist for unsupervised
  learning.
\newblock \emph{Phys. Rev. E}, 100:\penalty0 033311, Sep 2019.

\bibitem[Zhang and Lin(2018)]{doi:10.1098/rspa.2018.0305}
Sheng Zhang and Guang Lin.
\newblock Robust data-driven discovery of governing physical laws with error
  bars.
\newblock \emph{Proc Math Phys Eng Sci.}, 474\penalty0 (2217), 2018.

\bibitem[Iten et~al.(2020)Iten, Metger, Wilming, Del~Rio, and
  Renner]{iten2020discovering}
Raban Iten, Tony Metger, Henrik Wilming, L{\'\i}dia Del~Rio, and Renato Renner.
\newblock Discovering physical concepts with neural networks.
\newblock \emph{Physical review letters}, 124\penalty0 (1):\penalty0 010508,
  2020.

\bibitem[Cranmer et~al.(2020)Cranmer, Sanchez{-}Gonzalez, Battaglia, Xu,
  Cranmer, Spergel, and Ho]{DBLP:conf/nips/CranmerSBXCSH20}
Miles~D. Cranmer, Alvaro Sanchez{-}Gonzalez, Peter~W. Battaglia, Rui Xu, Kyle
  Cranmer, David~N. Spergel, and Shirley Ho.
\newblock Discovering symbolic models from deep learning with inductive biases.
\newblock In \emph{NeurIPS}, 2020.

\bibitem[Raissi et~al.(2020)Raissi, Yazdani, and Karniadakis]{Raissi20Fluid}
Maziar Raissi, Alireza Yazdani, and George~Em Karniadakis.
\newblock Hidden fluid mechanics: Learning velocity and pressure fields from
  flow visualizations.
\newblock \emph{Science}, 367\penalty0 (6481):\penalty0 1026--1030, 2020.

\bibitem[Raissi et~al.(2019)Raissi, Perdikaris, and
  Karniadakis]{RAISSI2019PhysicsInformedNN}
M.~Raissi, P.~Perdikaris, and G.E. Karniadakis.
\newblock Physics-informed neural networks: A deep learning framework for
  solving forward and inverse problems involving nonlinear partial differential
  equations.
\newblock \emph{Journal of Computational Physics}, 378:\penalty0 686--707,
  2019.

\bibitem[Liu and Tegmark(2021)]{Liu21AIPoincare}
Ziming Liu and Max Tegmark.
\newblock Machine learning conservation laws from trajectories.
\newblock \emph{Phys. Rev. Lett.}, 126:\penalty0 180604, May 2021.

\bibitem[Xue et~al.(2021)Xue, Nasim, Zhang, Fan, Zhang, and
  El{-}Azab]{nanovoid_tracking}
Yexiang Xue, Md. Nasim, Maosen Zhang, Cuncai Fan, Xinghang Zhang, and Anter
  El{-}Azab.
\newblock Physics knowledge discovery via neural differential equation
  embedding.
\newblock In \emph{{ECML/PKDD}}, pages 118--134, 2021.

\bibitem[Chen et~al.(2018)Chen, Rubanova, Bettencourt, and
  Duvenaud]{chen2018neural}
Ricky~TQ Chen, Yulia Rubanova, Jesse Bettencourt, and David~K Duvenaud.
\newblock Neural ordinary differential equations.
\newblock \emph{{NeurIPS}}, 31, 2018.

\bibitem[Valdés-Pérez(1994)]{Valdes1994}
R.E. Valdés-Pérez.
\newblock Human/computer interactive elucidation of reaction mechanisms:
  application to catalyzed hydrogenolysis of ethane.
\newblock \emph{Catalysis Letters}, 28:\penalty0 79--87, 1994.

\bibitem[King et~al.(2004)King, Whelan, Jones, Reiser, Bryant, Muggleton, Kell,
  and Oliver]{king2004functional}
Ross~D King, Kenneth~E Whelan, Ffion~M Jones, Philip~GK Reiser, Christopher~H
  Bryant, Stephen~H Muggleton, Douglas~B Kell, and Stephen~G Oliver.
\newblock Functional genomic hypothesis generation and experimentation by a
  robot scientist.
\newblock \emph{Nature}, 427\penalty0 (6971):\penalty0 247--252, 2004.

\bibitem[King et~al.(2009)King, Rowland, Oliver, Young, Aubrey, Byrne, Liakata,
  Markham, Pir, Soldatova, Sparkes, Whelan, and Clare]{king2009autosci}
Ross~D. King, Jem Rowland, Stephen~G. Oliver, Michael Young, Wayne Aubrey, Emma
  Byrne, Maria Liakata, Magdalena Markham, Pinar Pir, Larisa~N. Soldatova,
  Andrew Sparkes, Kenneth~E. Whelan, and Amanda Clare.
\newblock The automation of science.
\newblock \emph{Science}, 324\penalty0 (5923):\penalty0 85--89, 2009.

\bibitem[Biggio et~al.(2021)Biggio, Bendinelli, Neitz, Lucchi, and
  Parascandolo]{DBLP:conf/icml/BiggioBNLP21}
Luca Biggio, Tommaso Bendinelli, Alexander Neitz, Aur{\'{e}}lien Lucchi, and
  Giambattista Parascandolo.
\newblock Neural symbolic regression that scales.
\newblock In \emph{{ICML}}, volume 139, pages 936--945. {PMLR}, 2021.

\bibitem[Kamienny et~al.(2022)Kamienny, d'Ascoli, Lample, and
  Charton]{DBLP:conf/nips/KamiennydLC22}
Pierre{-}Alexandre Kamienny, St{\'{e}}phane d'Ascoli, Guillaume Lample, and
  Fran{\c{c}}ois Charton.
\newblock End-to-end symbolic regression with transformers.
\newblock In \emph{NeurIPS}, 2022.

\bibitem[Langley(1977)]{DBLP:conf/ijcai/Langley77}
Pat Langley.
\newblock {BACON:} {A} production system that discovers empirical laws.
\newblock In \emph{{IJCAI}}, page 344, 1977.

\bibitem[Langley(1979)]{DBLP:conf/ijcai/Langley79}
Pat Langley.
\newblock Rediscovering physics with {BACON.3}.
\newblock In \emph{{IJCAI}}, pages 505--507, 1979.

\bibitem[Langley et~al.(1981)Langley, Bradshaw, and
  Simon]{DBLP:conf/ijcai/LangleyBS81}
Pat Langley, Gary~L. Bradshaw, and Herbert~A. Simon.
\newblock {BACON.5:} the discovery of conservation laws.
\newblock In \emph{{IJCAI}}, pages 121--126, 1981.

\bibitem[Cerrato et~al.(2023)Cerrato, Brugger, Schmitt, and
  Kramer]{cerrato2023rlsci}
Mattia Cerrato, Jannis Brugger, Nicolas Schmitt, and Stefan Kramer.
\newblock Reinforcement learning for automated scientific discovery.
\newblock In \emph{AAAI Spring Symposium}, 2023.

\bibitem[Brence et~al.(2021)Brence, Todorovski, and
  Dzeroski]{DBLP:journals/kbs/BrenceTD21}
Jure Brence, Ljupco Todorovski, and Saso Dzeroski.
\newblock Probabilistic grammars for equation discovery.
\newblock \emph{Knowl. Based Syst.}, 224:\penalty0 107077, 2021.

\bibitem[Gec et~al.(2022)Gec, Omejc, Brence, Dzeroski, and
  Todorovski]{DBLP:conf/dis/GecOBDT22}
Bostjan Gec, Nina Omejc, Jure Brence, Saso Dzeroski, and Ljupco Todorovski.
\newblock Discovery of differential equations using probabilistic grammars.
\newblock In \emph{{DS}}, volume 13601, pages 22--31. Springer, 2022.

\bibitem[Dubc{\'{a}}kov{\'{a}}(2011)]{DBLP:journals/gpem/Dubcakova11}
Ren{\'{a}}ta Dubc{\'{a}}kov{\'{a}}.
\newblock Eureqa: software review.
\newblock \emph{Genet. Program. Evolvable Mach.}, 12\penalty0 (2):\penalty0
  173--178, 2011.

\bibitem[d'Ascoli et~al.(2024)d'Ascoli, Becker, Mathis, Schwaller, and
  Kilbertus]{2023odeformer}
St{\'e}phane d'Ascoli, S{\"o}ren Becker, Alexander Mathis, Philippe Schwaller,
  and Niki Kilbertus.
\newblock Odeformer: Symbolic regression of dynamical systems with
  transformers.
\newblock In \emph{{ICLR}}. OpenReview.net, 2024.

\bibitem[Koza(1994)]{koza1994genetic}
John~R Koza.
\newblock Genetic programming as a means for programming computers by natural
  selection.
\newblock \emph{Statistics and computing}, 4:\penalty0 87--112, 1994.

\bibitem[Ryan and Morgan(2007)]{doi:10.1080/15598608.2007.10411855}
Thomas~P. Ryan and J.~P. Morgan.
\newblock Modern experimental design.
\newblock \emph{Journal of Statistical Theory and Practice}, 1\penalty0
  (3-4):\penalty0 501--506, 2007.

\bibitem[Chen and Xue(2022)]{chen2022generalisation}
Qi~Chen and Bing Xue.
\newblock Generalisation in genetic programming for symbolic regression:
  Challenges and future directions.
\newblock In \emph{Women in Computational Intelligence: Key Advances and
  Perspectives on Emerging Topics}, pages 281--302. Springer, 2022.

\bibitem[Keren et~al.(2023)Keren, Liberzon, and
  Lazebnik]{keren2023computational}
Liron~Simon Keren, Alex Liberzon, and Teddy Lazebnik.
\newblock A computational framework for physics-informed symbolic regression
  with straightforward integration of domain knowledge.
\newblock \emph{Scientific Reports}, 13\penalty0 (1):\penalty0 1249, 2023.

\bibitem[Haut et~al.(2022)Haut, Banzhaf, and Punch]{DBLP:conf/gecco/HautBP22}
Nathan Haut, Wolfgang Banzhaf, and Bill Punch.
\newblock Active learning improves performance on symbolic regression tasks in
  stackgp.
\newblock In \emph{{GECCO} Companion}, pages 550--553, 2022.

\bibitem[Haut et~al.(2023)Haut, Punch, and Banzhaf]{DBLP:conf/gecco/HautPB23}
Nathan Haut, Bill Punch, and Wolfgang Banzhaf.
\newblock Active learning informs symbolic regression model development in
  genetic programming.
\newblock In \emph{{GECCO} Companion}, pages 587--590, 2023.

\bibitem[Nagelkerke et~al.(1991)]{nagelkerke1991note}
Nico~JD Nagelkerke et~al.
\newblock A note on a general definition of the coefficient of determination.
\newblock \emph{Biometrika}, 78\penalty0 (3):\penalty0 691--692, 1991.

\bibitem[Brechmann and Rendall(2021)]{brechmann2021unbounded}
Pia Brechmann and Alan~D Rendall.
\newblock Unbounded solutions of models for glycolysis.
\newblock \emph{Journal of mathematical biology}, 82:\penalty0 1--23, 2021.

\bibitem[De~Bartolo and Carbone(2006)]{de2006role}
R~De~Bartolo and Vincenzo Carbone.
\newblock The role of the basic three-modes interaction during the free decay
  of magnetohydrodynamic turbulence.
\newblock \emph{Europhysics Letters}, 73\penalty0 (4):\penalty0 547, 2006.

\end{thebibliography}
